\documentclass{article}
\usepackage{amssymb}

\usepackage[final]{corl_2025} 

\usepackage[numbers]{natbib}
\usepackage{multicol}
\usepackage{graphicx}
\usepackage{amsmath, bm}
\usepackage{multirow}
\usepackage{booktabs}
\usepackage[table]{xcolor}
\usepackage{caption}
\usepackage{titlesec}
\usepackage{setspace}
\usepackage{titletoc}

\title{Neural Robot Dynamics}

\newcommand{\eg}{\textit{e.g.},~}
\newcommand{\ie}{\textit{i.e.},~}

\hypersetup{
    colorlinks=true,
    urlcolor=magenta,
    pdftitle={Neural Robot Dynamics},
}
    
\author{
  Jie Xu$^{1}$, Eric Heiden$^{1}$, Iretiayo Akinola$^{1}$, Dieter Fox$^{1,2}$, Miles Macklin$^{1}$, Yashraj Narang$^{1}$\\
  $^{1}$NVIDIA\,\,\,\, $^{2}$University of Washington\\
  \href{https://neural-robot-dynamics.github.io/}{https://neural-robot-dynamics.github.io}
}

\begin{document}
\maketitle

\setlength{\abovedisplayskip}{4pt} 
\setlength{\belowdisplayskip}{4pt}

\begin{abstract}
Accurate and efficient simulation of modern robots remains challenging due to their high degrees of freedom and intricate mechanisms. 
Neural simulators have emerged as a promising alternative to traditional analytical simulators, capable of efficiently predicting complex dynamics and adapting to real-world data; however, existing neural simulators typically require application-specific training and fail to generalize to novel tasks and/or environments, primarily due to inadequate representations of the global state. 
In this work, we address the problem of learning generalizable neural simulators for robots that are structured as articulated rigid bodies. We propose \textit{NeRD} (Neural Robot Dynamics), learned robot-specific dynamics models for predicting future states for articulated rigid bodies under contact constraints. \textit{NeRD} uniquely replaces the low-level dynamics and contact solvers in an analytical simulator and employs a robot-centric and spatially-invariant simulation state representation. We integrate the learned \textit{NeRD} models as an interchangeable backend solver within a state-of-the-art robotics simulator. We conduct extensive experiments to show that the \textit{NeRD} simulators are stable and accurate over a thousand simulation steps; generalize across tasks and environment configurations; enable policy learning exclusively in a neural engine; and, unlike most classical simulators, can be fine-tuned from real-world data to bridge the gap between simulation and reality.

\end{abstract}


    
\section{Introduction}

Simulation plays a crucial role in various robotics applications, such as policy learning \cite{andrychowicz2020learning, chen2023visualdex, Haarnoja2024learning, handa2023dextreme, kumar2021rma, lee2020learning, tang2024automate}, safe and scalable robotic control evaluation \cite{funk2021benchmarking,gu2023maniskill,li24simpler,liu2021ocrtoc}, and computational optimization of robot designs \cite{du2016_copter, li2023tool, xu2021design}. 
Recently, neural robotics simulators have emerged as a promising alternative to traditional analytical simulators, as neural simulators can efficiently predict robot dynamics and learn intricate physics from real-world data.
For instance, neural simulators have been leveraged to capture complex interactions challenging for analytical modeling \cite{pfrommer2021contactnets, hoffman2025learning, ajay2018augmenting, zeng2020tossingbot}, or have served as learned world models to facilitate sample-efficient policy learning \cite{janner2019mbpo,li2025roboticworldmodelneural}. 

However, existing neural robotics simulators typically require application-specific training, often assuming fixed environments \cite{li2025roboticworldmodelneural,andriluka2024neural} or simultaneous training alongside control policies \cite{fussell2021supertrack, hansen2022temporal}. 
These limitations primarily stem from their end-to-end frameworks with inadequate representations of the global simulation state, \ie neural models often substitute the entire classical simulator and directly map robot state and control actions (\eg target joint positions, target link orientations) to the robot's next state. 
Without encoding the environment in the state representation, the learned simulators have to implicitly memorize the task and environment details. Additionally, utilizing controller actions as input causes the simulators to overfit to particular low-level controllers used during training. Consequently, unlike classical simulators, these neural simulators often fail to generalize to novel state distributions (induced by new tasks), unseen environment setups, and customized controllers (\eg novel control laws or controller gains).

\begin{figure*}[t!]
    \centering
    \includegraphics[width=\linewidth]{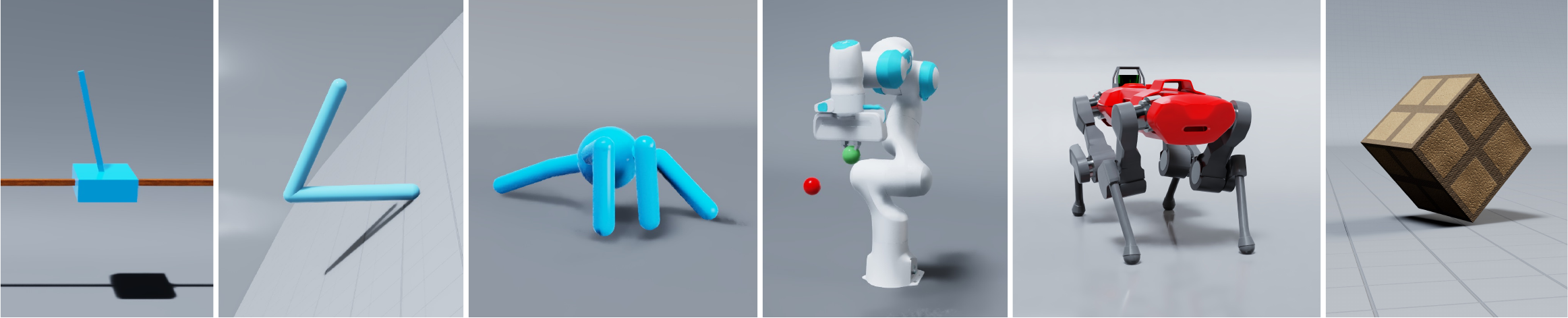}

    \caption{We propose \textbf{\textit{NeRD}}, learned robot-specific dynamics models for generalizable articulated rigid body simulation. We demonstrate our approach by training \textit{NeRD} models on six diverse robotic systems, from left: \textit{Cartpole, Double Pendulum, Ant, Franka, ANYmal, Cube Toss}.}
    \label{fig:tasks}
    \vspace{-1em}
\end{figure*}

In this work, we address the problem of learning generalizable neural simulators for articulated rigid-body robots. We envision a future where each robot is equipped with a neural simulator pretrained from analytical simulations. Such a simulator could conduct lifelong fine-tuning as the robot interacts with the real environment to accommodate wear-and-tear and environmental changes, and facilitate versatile skill learning in a digital twin powered by the continuously-updated simulator. 

Toward this goal, we propose \textit{Neural Robot Dynamics} (\textit{NeRD}), a learned robot-specific dynamics model for predicting the evolution of articulated rigid-body states under contact constraints. \textit{NeRD} is characterized by two key innovations: (1) a \textit{hybrid prediction framework},  where \textit{NeRD} uniquely replaces only the \textit{application-agnostic} simulation modules -- \ie the low-level forward dynamics and contact solvers -- and leverages a
general and compact 
representation describing the world surrounding robots;
(2) a \textit{robot-centric state representation}, 
where \textit{NeRD} further improves the simulation state representation 
to explicitly enforce dynamics invariance under translation and rotation around the gravity axis, thus further enhancing \textit{NeRD}'s spatial generalizability and training efficiency. 
Once trained, our \textit{NeRD} models can (1) provide stable and accurate predictions over hundreds to thousands of simulation steps; (2) generalize to different tasks, environments, and low-level controllers; and (3) effectively fine-tune from real-world data to bridge sim-to-real gaps. Additionally, with our 
hybrid and modular 
design of \textit{NeRD}, we integrate \textit{NeRD} as an interchangeable backend solver within a state-of-the-art robotics simulator \cite{warp2022}, enabling users to effortlessly reuse existing policy-learning environments and activate \textit{NeRD} as a new physics backend through a single-line switch. 

We train \textit{NeRD} models on six different robotic systems (Fig.~\ref{fig:tasks}) to illustrate the broad applicability of our proposed methodology. We evaluate the trained \textit{NeRD} models on extensive experiments in both simulation and real-world scenarios, including long-horizon dynamics prediction and policy learning on a diverse set of tasks that are unseen during \textit{NeRD} model training. 
Due to the long-horizon stability and accuracy of \textit{NeRD}, we demonstrate -- for the first time, to the best of our knowledge -- that robotic policies learned exclusively within a pretrained neural simulator can successfully achieve zero-shot deployment in the analytical simulator and even transfer directly to the real world.

\section{Related Work}
Neural physics simulations have been studied across diverse simulation domains, including cloth~\citep{bertiche2022cloth, jin2024neural, pfaff2021learning}, fluid~\citep{ladicky2015data,sanchezgonzalez2020learning}, and continuum dynamics~\citep{li2024mpmnet, li2023pac}.
Our work focuses on the subfield of neural simulation for articulated rigid-body dynamics in robotics. 

Neural physics engines for single rigid bodies have modeled object-ground and inter-object interactions~\cite{pfrommer2021contactnets, jiang2022contact, allen2023contact, allen2023learning}. ContactNets~\cite{pfrommer2021contactnets} learns implicit signed distance functions to capture the discontinuous cube-ground dynamics, while a subsequent method~\cite{allen2023contact} employs graph neural networks (GNNs) to improve the prediction accuracy of the same task. \citet{allen2023learning} further model inter-object collisions with face interaction graph networks. Despite their advancements, these approaches are not readily extendable to articulated rigid bodies, limiting practicality in robotics applications. 

Neural models predicting dynamics of articulated rigid bodies have been explored in model-based reinforcement learning and planning, known as world models. Most model-based RL methods~\cite{janner2019mbpo, li2025roboticworldmodelneural, fussell2021supertrack, hansen2022temporal, Hafner2020Dream, hafner2023dreamerv3, hansen2024tdmpc2, moerland2023mbrl} predict future robot states directly from the current robot state and control actions, without explicit environment modeling, and jointly train the neural world models with control policies. Consequently, these world models lack generalizability to novel tasks, environments, and controllers. 
Some works in model-based planning decouple the simulation model training from planning. For instance, GNNs~\cite{sanchez2018graph} have been utilized to model generalizable physics across articulated rigid bodies. But this approach still directly predicts state transition from robot state and action, primarily targeting 2D systems or contact-free dynamics. A concurrent work~\cite{hoffman2025learning} pretrains a Bayesian network for modeling the dynamics of a loco-manipulation system but relies on analytical modeling for this particular robot, restricting its applicability to broader robot systems.

A recent work, LARP~\cite{andriluka2024neural}, couples dynamics and contact networks for modeling humanoid-ball interactions, but targets human motion reconstruction in computer vision, with the accuracy and applicability to robotic policy learning unverified. Physics-informed neural networks~\cite{raissi2019pinn, cuomo2022pinn} incorporate physics laws into learning generalizable dynamics of simple articulations, but their reliance on expert-modeled physics laws of each individual system limits their use in complex robot designs.

Hybrid neural simulation frameworks have also been proposed. NeuralSim~\cite{heiden2021neuralsim} integrates neural models in localized components of a rigid-body simulator to improve friction and passive force modeling. But models' generalizability is limited by robot-state-only representations. Neural contact clustering~\cite{kim2019contact} and neural collision detectors~\cite{son2023local} accelerate contact algorithms. Residual physics is also studied~\cite{ajay2018augmenting, zeng2020tossingbot} for bridging sim-to-real gaps. These techniques complement ours, as collision detection generates contact information consumed by \textit{NeRD}, and the residual physics augments \textit{outputs} of an analytical simulator while we propose a generalizable \textit{input} for a neural simulator.

\section{NeRD: Neural Robot Dynamics}

We now present \textit{NeRD}, robot-specific neural dynamics models for articulated rigid-body simulation. We start by presenting the typical workflow of a classical articulated rigid-body simulator in \S\ref{sec:classical_sim}. Next,  in \S\ref{sec:nerd_framework}, we introduce the \textit{hybrid prediction framework} of \textit{NeRD}, which leverages a general and compact simulation state representation describing the world to enable generalization across applications. In \S\ref{sec:robot-centric_representation}, we further improve the representation by proposing a \textit{robot-centric simulation state representation} to enforce spatial generalizability and improve the training efficiency of \textit{NeRD}.

\subsection{Preliminary: A Typical Robotics Simulation Workflow}
\label{sec:classical_sim}
\begin{figure*}[t!]
    \centering
    
    \includegraphics[width=\linewidth]{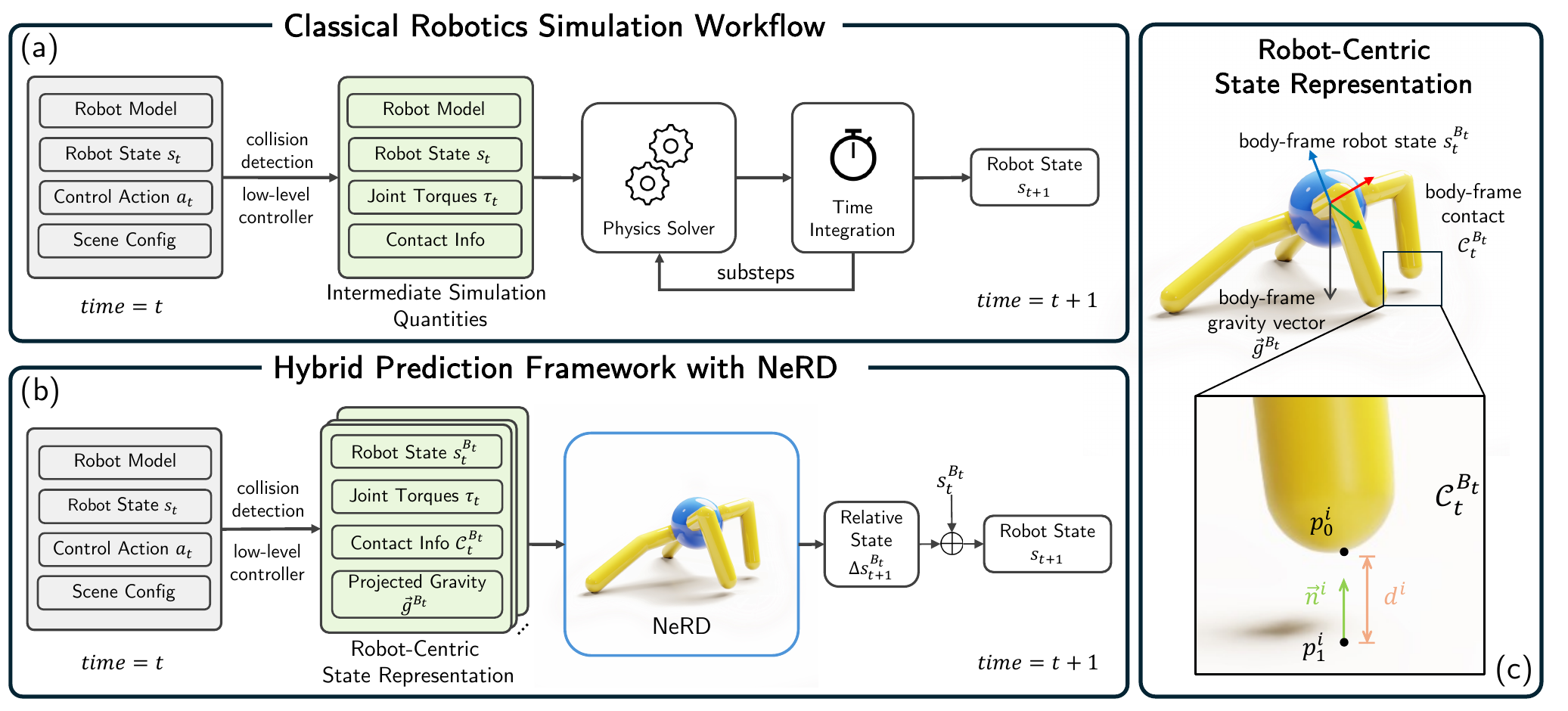}
    \caption{\textbf{Framework overview for Neural Robot Dynamics (\textit{NeRD}).} \textbf{(a)} Workflow of a classical robotics simulator. The quantities shaded in green are application-agnostic. \textbf{(b)} Hybrid prediction framework of the \textit{NeRD}-integrated simulator. Inputs to \textit{NeRD} are the robot-centric state representations (illustrated in (\textbf{c})) within a history window.}
    \label{fig:overview}
\end{figure*}

We illustrate a typical workflow of a classical robotics simulator in Fig. \ref{fig:overview}(a). The user first sets up the simulator by importing a robot model with its initial state, and specifying an environment configuration (\eg ground, objects) and a low-level controller (\eg joint position control, end-effector control). At each time step $t$, the simulator takes as input the robot model, current robot state $\bm{s}_t$, the action command fed to the robot $\bm{a}_t$, and the scene configuration. It then performs collision detection to identify contact information for interacting physical parts, and executes the low-level controller to convert the action command into joint-space torques. Those, along with the robot state, serve as intermediate quantities that are processed by the dynamics and contact solvers, where physics equations are formulated and a numerical solver is employed to calculate the acceleration. Finally, the simulator performs time integration to obtain the new state of the robot. 

Previous neural robotics simulators \cite{janner2019mbpo, li2025roboticworldmodelneural, fussell2021supertrack, hansen2024tdmpc2} often adopt an \textit{end-to-end} framework that substitutes the entire simulation engine with a neural model and directly maps the robot state $\bm{s}_t$ and action command $\bm{a}_t$ to the next robot state $\bm{s}_{t+1}$, without leveraging information regarding the scene and the controller, \ie $\textit{E2E}(\bm{s}_t, \bm{a}_t)\rightarrow\bm{s}_{t+1}$. Such neural simulators are therefore forced to memorize the scene and the controller used for training and lack generalizability to novel applications.

\subsection{Hybrid Prediction Framework of Neural Robot Dynamics}
\label{sec:nerd_framework}
To train a generalizable neural simulator, we need a comprehensive representation to encode the scene and controller that generalizes across diverse applications. 
Inspired by the observation that the low-level dynamics and contact solvers in a classical simulator are application-agnostic, \textit{NeRD} employs a \textit{hybrid prediction framework} that replaces only the core physics components in a conventional simulator (Fig.~\ref{fig:overview}(b)). This hybrid framework allows \textit{NeRD} to leverage intermediate simulation quantities (\ie robot state, contact information, and joint-space torques) as a general and compact representation describing the full simulation state, providing all necessary information to evolve the robot dynamics regardless of the applications (\eg tasks, scenes, and controllers).

Formally speaking, let $\bm{s}_t = (\bm{x}_t, \bm{R}_t, \bm{q}_t, \bm{\phi}_t, \dot{\bm{q}}_t)$ denote the robot state at time $t$, where $\bm{x}_t$ and $\bm{R}_t$ are the position and orientation (represented as a quaternion) of the robot base, $\bm{q}_t$ denotes articulated joint angles, $\bm{\phi}_t$ is the spatial twist of the base (\ie 6D velocity), and $\dot{\bm{q}}_t$ are joint velocities. We define $\bm{\tau}_t$ as the joint-space torque and $\bm{\mathcal{C}}_t$ as contact-related quantities. We construct $\bm{\mathcal{C}}_t = \{\bm{c}_t^i\}$ by reusing the collision detection module in the classical simulator. For each pre-specified contact point $\bm{p}_0^i$ on the robot, we obtain its contact event quantities $\bm{c}_t^i = (\bm{p}_0^i, \bm{p}_1^i, \vec{\bm{n}}^i, d^i)$. Here $\bm{p}_1^i$ is the contact point on a non-robot shape, $\vec{\bm{n}}^i$ is the contact normal, and $d^i$ is the contact distance.

Our neural robot dynamics model is a parametric function $\textit{NeRD}_\theta\left(\{\bm{s}_k, \bm{\mathcal{C}}_k, \bm{\tau}_k\}_{k=t-h+1}^t\right)$ that maps the robot states, contacts, and joint torques within a history window of length $h$ to the state difference $\Delta \bm{s}_{t+1}\triangleq\bm{s}_{t+1} \ominus \bm{s}_{t}$; here, $\ominus$ is defined to be the rotation difference $\textbf{R}_{t+1}\textbf{R}_t^{-1}$ for the base orientation and the subtraction operator for other state dimensions. The model is trained by minimizing the mean squared error between the prediction and the ground-truth state difference $\widehat{\Delta \bm{s}_{t+1}}$:
\begin{equation}
\label{eq:original_loss}
    \mathcal{L}_\theta = \frac{1}{NS}\sum_N \|\textit{NeRD}_\theta\left(\{\bm{s}_k, \bm{\mathcal{C}}_k, \bm{\tau}_k\}_{k=t-h+1}^t\right) - \widehat{\Delta \bm{s}_{t+1}}\|^2,
\end{equation}
where $N$ is the batch size and $S$ is the dimension of the robot state. The next state $\bm{s}_{t+1}$ is then computed by  $\bm{s}_{t+1}=\bm{s}_{t} \oplus \textit{NeRD}_\theta\left(\{\bm{s}_k, \bm{\mathcal{C}}_k, \bm{\tau}_k\}_{k=t-h+1}^t\right)$. 
This concise state representation $\{\bm{s}_t, \bm{\mathcal{C}}_t, \bm{\tau}_t\}$ is a carefully designed outcome resulting from deeply integrating the neural models into the classical simulation framework and reusing the application-agnostic intermediate simulation quantities, thereby fundamentally providing the generalizability across diverse applications.

\subsection{Robot-Centric State Representation}
\label{sec:robot-centric_representation}
The dynamics of a robot remain invariant under spatial translation, as well as rotation around the gravity axis, provided that its interaction with the environment, such as contact forces, remains unchanged in the robot's body frame. Inspired by this, we enhance the simulation state representation by introducing a \textit{robot-centric parameterization} to explicitly enforce such spatial invariance.

Specifically, we transform the robot state $\bm{s}_t$ and contact-related quantities $\bm{\mathcal{C}}_t$ into the robot's base frame $\bm{B}_t = (\bm{x}_t, \bm{R}_t)$, as shown in Fig.~\ref{fig:overview}(c). For the robot articulation, we use the reduced coordinate state, which is spatially invariant; thus, we only need to transform the state of the robot base (\ie $\bm{x}_t$, $\bm{R}_t$, and $\bm{\phi}_t$) into the robot's base frame. To properly account for gravity when the robot rotates about axes other than the gravity axis, we treat gravity as an external force and augment the simulation state with gravity expressed in the robot's base frame. Additionally, the predicted state difference $\Delta \bm{s}_{t+1}$ (\ie network's output) is also expressed in the robot's base frame $\bm{B}_t$. Intuitively, this robot-centric representation encodes the world from the robot's local, myopic view -- knowing its joint state, how it contacts the environment locally, and how external forces (\ie gravity) are applied to it. \textit{NeRD} then uses this information to evolve the robot's dynamics within the local frame. By using this robot-centric parameterization, we reformulate our loss function from Eq. \ref{eq:original_loss} as:
\begin{equation}
\label{eq:loss}
\mathcal{L}_\theta = \frac{1}{NS} \sum_{N}  \Big\| \textit{NeRD}_\theta \Big( \{ \bm{s}_k^{\bm{B}_k}, \bm{\mathcal{C}}_k^{\bm{B}_k}, 
\bm{\tau}_{k}, \vec{\bm{g}}^{\bm{B}_k} \}_{k=t-h+1}^{t} \Big) - \widehat{\Delta \bm{s}_{t+1}^{\bm{B}_{t}}} \Big\|,
\end{equation}
where $\vec{\bm{g}}$ is the unit gravity vector, and the superscript $\bm{B}_k$ (or $\bm{B}_t$) means the corresponding quantity is expressed in the robot base frame at timestep $k$ (or $t$). 
The robot state is then updated by 
\begin{equation}
\bm{s}_{t+1} = \mathcal{T}^w_{\bm{B}_t}\Big(\bm{s}^{\bm{B}_t}_{t} \oplus \textit{NeRD}_\theta\left(\{\bm{s}_k^{\bm{B}_k}, \bm{\mathcal{C}}_k^{\bm{B}_k}, \bm{\tau}_{k}, \vec{\bm{g}}^{\bm{B}_k}\}_{k=t-h+1}^t\right)\Big),
\end{equation}
where $\mathcal{T}^w_{\bm{B}_t}(\cdot)$ is the transformation from robot base frame at time step $t$ to the world frame.

The spatial invariance property of our robot-centric representation explicitly enforces the spatial generalizability of the learned robot dynamics models.
In addition, it reduces the state space, substantially enhancing the training and data efficiency of \textit{NeRD} by eliminating the need to exhaustively sample all spatial positions and orientations of the robot during model training. 

\section{Implementation}
\textit{NeRD} is compatible with most articulated rigid-body simulation frameworks, as it uses intermediate quantities commonly computed in a standard simulator. We validate our approach by integrating \textit{NeRD} into a state-of-the-art robotics simulator, NVIDIA's Warp simulator \cite{warp2022}, since Warp's modular design enables implementing \textit{NeRD} as an interchangeable solver module in Python and keeps it transparent to simulation users. We use a GPU-parallelized collision detection algorithm adapted from the one in Warp. 

\textbf{Training Datasets}\quad 
We generate the training datasets for \textit{NeRD} in a task-agnostic manner using Warp with the Featherstone solver \cite{feathersone2007}. For each robot instance in our experiments, we collect $100$K random trajectories, each consisting of $100$ timesteps. These trajectories are generated using randomized initial states of the robot, random joint torque sequences within the robot's motor torque limits, and optionally, randomized environment configurations. 


\textbf{Network and Training Details}\quad 
We model \textit{NeRD} using a causal Transformer architecture, specifically a lightweight implementation of the GPT-2 Transformer \cite{nanoGPT,radford2019language}.
We use a history window size $h = 10$ for all tasks in our experiments. During training, we sample batches of sub-trajectories of length $h$ and train the model using a teacher-forcing approach \cite{Ilya2014sequence}. 
To prevent the loss from being dominated by high-variance velocity terms, 
we normalize the output prediction, using the mean and standard deviation statistics computed from the dataset. 
Ablation experiments (see Appendix~\ref{sup:ablation}) show that output normalization is critical for improving the accuracy and long-horizon stability of \textit{NeRD}. 
We also apply normalizations to the model's input to regularize the ranges of the inputs, improving the stability of model training. We report training hyperparameters in the Appendix~\ref{sup:training_details}.

\section{Experiments}
\label{sec:experiments}

We train \textit{NeRD} models on six distinct robotic systems (Fig. \ref{fig:tasks}) and conduct extensive experiments to validate the capabilities of the trained \textit{NeRD} models \footnote{View the supplementary video to observe qualitative performance throughout the following examples.}. We investigate the following questions: Can \textit{NeRD} reliably and accurately simulate long-horizon robotic trajectories (\S\ref{sec:exp-passive})? Does \textit{NeRD}'s hybrid prediction framework enable it to generalize across different contact configurations (\S\ref{sec:exp-contact})?  Can a single \textit{NeRD} model generalize to diverse tasks, customized robotic controllers, and spatial regions that are unseen during training? Can we train robotic policies for diverse tasks entirely in a \textit{NeRD} simulator and successfully deploy the learned policies in the ground-truth simulator and even in the real world (\S\ref{sec:exp-policy-learning}, \S\ref{sec:exp-sim-to-real})? Finally, can we effectively fine-tune the pretrained \textit{NeRD} models from real-world data (\S\ref{sec:exp-cube-tossing})? We also provide a comprehensive ablation study in Appendix~\ref{sup:ablation}, highlighting critical design decisions essential for successfully training \textit{NeRD} models.

\subsection{Long-Horizon Stability and Accuracy: Cartpole and Ant with Passive Motion}
\label{sec:exp-passive}
\begin{figure}[t!]
    \begin{minipage}[c]{0.59\textwidth}
        \vspace{0pt}
        \centering
        \resizebox{\textwidth}{!}{

            \begin{tabular}{lcccc}
                \toprule
                \textbf{Robot} & \multicolumn{3}{c}{\textbf{Cartpole}} & \textbf{Ant} \\ 
                \cmidrule(lr){2-4}\cmidrule(lr){5-5}
                \textbf{Trajectory Horizon} & 100 & 500 & 1000 & 500 \\ 
                \midrule
                Base Position Err. (\textit{m}) & $0.0002$ & $0.004$ & $0.033$ & $0.057$ \\
                Base Orientation Err. (\textit{rad}) & - & - & - & $0.095$\\
                Non-Base Joint Err. (\textit{rad}) & $0.0004$ & $0.013$ & $0.075$ & $0.077$ \\
                \bottomrule
            \end{tabular}
        }
    \end{minipage} 
    \begin{minipage}[c]{0.45\textwidth}
        \vspace{0pt}
        \centering
        \includegraphics[width=0.8\textwidth]{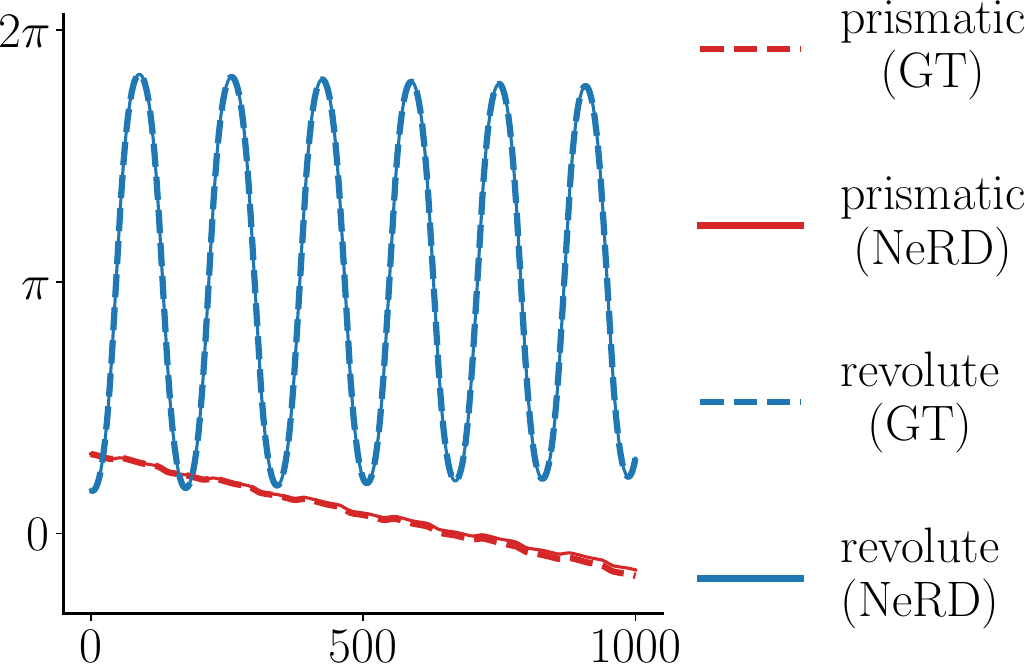}
    \end{minipage}
\captionof{figure}{\textbf{Evaluation of \textit{NeRD} on long-horizon passive motions.} Left: Full report of the measured errors. Right: $1000$-step cartpole state trajectories simulated by \textit{NeRD} and ground-truth simulator. }
\label{fig:passive_motion_eval}
\end{figure}

We first evaluate \textit{NeRD}'s long-horizon performance using open-loop passive motions of \textit{Cartpole} and \textit{Ant}. While \textit{Cartpole} is a contact-free system that serves as a tractable problem for analysis, \textit{Ant} assesses \textit{NeRD}'s performance when a floating base, high DoFs ($14$), and contact are all involved. The robots start from randomized initial states and apply zero joint torques, $\bm{\tau}_t=0$. We compute the temporally averaged errors between the trajectories generated by \textit{NeRD} and the ground-truth simulators. We report the errors averaged from $2048$ trajectories for each test in Fig.~\ref{fig:passive_motion_eval} (left).  

For \textit{Cartpole}, we evaluate trajectories of $100$, $500$, and $1000$ steps. We measure the errors of the prismatic base joint (reported as \textit{Base Position Err.}) and the non-base revolute joint. As a reference for the prismatic joint error, the pole length is $1$ \textit{m}. 
To visualize the simulation accuracy, Fig.~\ref{fig:passive_motion_eval} (right) compares the state trajectories generated by \textit{NeRD} and the ground-truth simulator from the same initial state. 
The results demonstrate high long-horizon accuracy of \textit{NeRD} on \textit{Cartpole}, with accumulated error of $0.075$ \textit{rad} (smaller than $5^\circ$) for the revolute joint and $0.033$ \textit{m} for the prismatic joint, even after $1000$ steps (\ie equivalent to $16.67$ seconds of passive motion). 
For \textit{Ant}, we evaluate on $500$-step trajectories, as the motion typically converges to a static state within this duration. We measure the base's position and orientation errors, and the mean error of non-base revolute joints. \textit{NeRD} achieves an average base angular error of  $0.095$ \textit{rad} and positional error of $0.057$ \textit{m} after $500$ steps of simulation ($1.2\,m$ full body width). The minimal prediction errors indicate \textit{NeRD}'s capability to accurately predict the motion of robots with a floating base for extended horizons.

\subsection{Contact Generalizability: Double Pendulum with Varying Contact Environments}
\label{sec:exp-contact}

We validate \textit{NeRD}'s generalizability across varying contact configurations using a \textit{Double Pendulum} example, in which a randomized planar ground (random normal direction and position) is placed beneath the double pendulum. Different combinations of ground configurations and initial pendulum states yield distinct modes of motion: \textit{contact-free chaotic motion} \cite{shinbrot1992chaos} when the ground is distant; \textit{sliding contact motion}, occurring when the pendulum lightly touches and slides along the ground surface; and \textit{collision-induced stopping motion} when the ground is positioned closely enough that the pendulum rapidly comes to rest after contact. Such varied contact configurations pose challenges for prior methods, as typical state representation without encoding the environment provides insufficient clues to determine contact timing and mode. To test \textit{NeRD}, we evaluate seven different ground setups -- one contact-free and six involving potential pendulum-ground contact. For each ground configuration, $2048$ passive-motion trajectories of $100$ steps are simulated with random initial states of the pendulum and zero joint torques.
Due to space constraints,  visualizations of the seven ground configurations and detailed error metrics for \textit{NeRD} are provided in the Appendix~\ref{sup:pendulum_results}.
Among all seven ground configurations, the maximum mean joint error after $100$-step simulation is $0.056$ \textit{rad} ($3.2^\circ$), with joint errors typically below $1^\circ$ for most cases. The results demonstrate that a single \textit{NeRD} model effectively generalizes across diverse contact scenarios.

    

\subsection{Task, Controller, and Spatial Generalizability: Robotic Policy Learning via RL}
\label{sec:exp-policy-learning}
To evaluate the task, controller, and spatial generalizability of \textit{NeRD}, we conduct extensive RL policy-learning experiments across diverse tasks for four robotic systems: a swing-up task for \textit{Cartpole}, an end-effector reach task for \textit{Franka}, three different tasks (running, spinning, and spin tracking) for \textit{Ant}, and forward and sideways velocity-tracking for \textit{ANYmal}~\cite{hutter2016anymal}. While detailed task descriptions are provided in the Appendix~\ref{sup:tasks}, we highlight key aspects here. First, each task explores specialized robot state distributions that is never covered in the \textit{NeRD} training dataset for that robot, which \textit{only} includes randomly-generated motions. Conducting policy learning in those tasks requires the trained \textit{NeRD} models to make accurate predictions under unseen state distributions. Second, to verify trained \textit{NeRD} models' generalizability to low-level controllers, we use \textit{joint-torque control} for \textit{Cartpole} and \textit{Ant}, and \textit{joint-position control} for \textit{Franka} and \textit{ANYmal}. Third, in the \textit{Ant} running task and \textit{ANYmal} velocity-tracking tasks, the robots reach a spatial region that is exceptionally far from the range covered by the training datasets, thus examining \textit{NeRD}'s spatial generalizability. Fourth, the horizons of the tasks vary from several hundred steps to $1000$ (\textit{ANYmal} tasks), assessing the stability and accuracy of the \textit{NeRD} models over extremely long horizons. 

For each task, we use PPO \cite{schulman2017proximal} to train three policies with different random seeds \textit{entirely} within the \textit{NeRD} simulators. We then evaluate each learned policy over $2048$ trajectories in \textit{both} the \textit{NeRD} and the ground-truth simulators, and report the average rewards and the standard deviations in Table~\ref{tab:policy_eval}. Despite training purely from random trajectories, the results show that the trained \textit{NeRD} models can support high-performing policy learning for diverse tasks (see supplementary video for policy behaviors). Furthermore, the \textit{NeRD}-trained policies have remarkably similar rewards when deployed in the \textit{NeRD} simulator and in the ground-truth simulator (without any fine-tuning or adaptation phase), further confirming the long-horizon predictive accuracy of \textit{NeRD} models.


\begin{table*}[t!]
\caption{Quantitative evaluation of policies trained exclusively in \textit{NeRD} simulators, when deployed in both \textit{NeRD} simulators \textit{and} the ground-truth simulator.}
\centering
\label{tab:policy_eval}
{
\fontsize{6.8pt}{7.5pt}\selectfont
\setlength{\tabcolsep}{4pt}
\begin{tabular}{lccccccc}
\toprule
\textbf{Robot}    & \textbf{Cartpole}         & \textbf{Franka}       & \multicolumn{3}{c}{\textbf{Ant}}                                    & \multicolumn{2}{c}{\textbf{ANYmal}} \\ 
\cmidrule(lr){2-2}\cmidrule(lr){3-3}\cmidrule(lr){4-6}\cmidrule(lr){7-8}
\textbf{Task}     & Swing Up                  & Reach                 & Running               & Spinning            & Spin Tracking         & Forward Walk       & Sideways Walk \\
\midrule
GT Reward         & 1212.5 $\pm$ 210.4        & 89.3 $\pm$ 10.5      & 2541.5 $\pm$ 309.1    & 2624.7 $\pm$ 641.0  & 1630.2 $\pm$ 203.1    & 1323.4 $\pm$ 60.5  & 1360.2 $\pm$ 81.2  \\
\cmidrule(lr){2-2}\cmidrule(lr){3-3}\cmidrule(lr){4-6}\cmidrule(lr){7-8}
NeRD Reward       & 1212.6 $\pm$ 210.2        & 91.1 $\pm$ 9.9      & 2649.5 $\pm$ 227.4    & 3076.2 $\pm$ 433.5  & 1670.5 $\pm$ 192.6    & 1323.1 $\pm$ 62.4  & 1359.2 $\pm$ 70.4  \\
\cmidrule(lr){2-2}\cmidrule(lr){3-3}\cmidrule(lr){4-6}\cmidrule(lr){7-8}
Reward Err. (\%) & +0.01\%                   & +2.02\%               & +4.25\%               & +17.21\%            & +2.47\%               & -0.02\%            & -0.07\%            \\
\bottomrule
\end{tabular}
}
\end{table*}

\subsection{Sim-to-Real Transfer of Franka Reach Policy}
\label{sec:exp-sim-to-real}
\begin{figure}[t!]
    \centering
    \includegraphics[width= 0.95\linewidth]{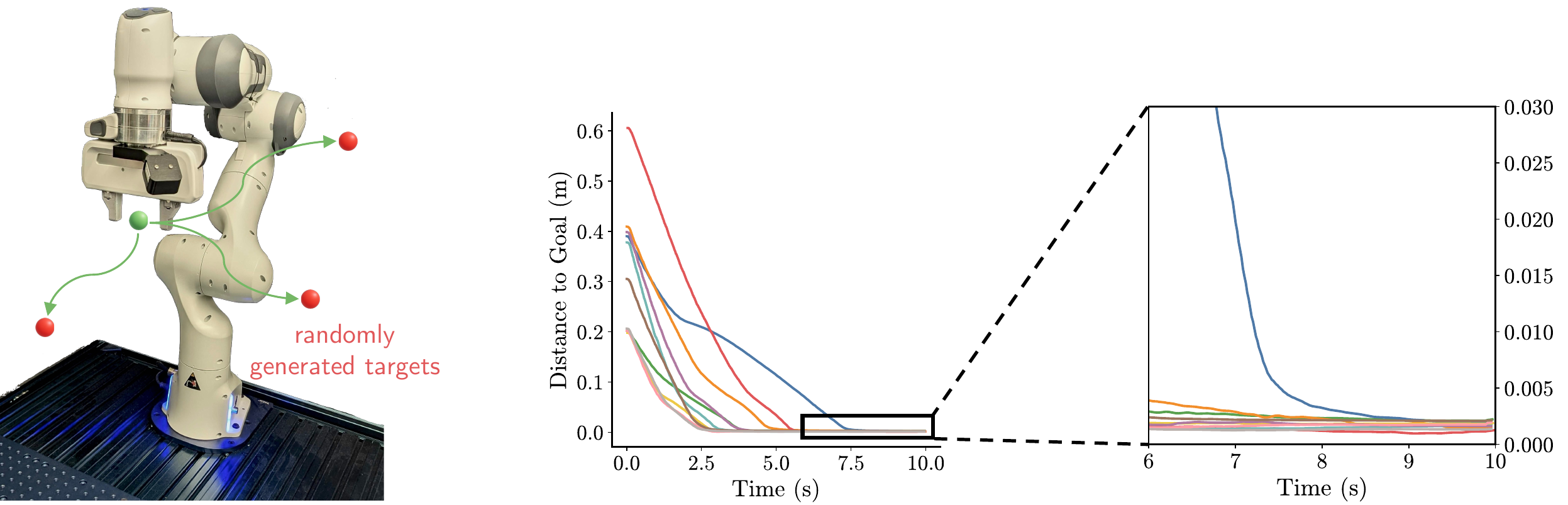}

    \caption{\textbf{Zero-shot sim-to-real transfer of a Franka reach policy.} The real-world setup is shown in the \textbf{left} figure. The plot on the \textbf{middle} visualizes the evolution of distance-to-goal measurements when $10$ \textit{NeRD}-trained policies are executed, with a zoomed-in plot in the \textbf{right}.}
    \label{fig:franka_reach}
\end{figure}

We further evaluate the accuracy of \textit{NeRD} models via zero-shot sim-to-real transfer of a \textit{Franka} reach policy trained exclusively in the \textit{NeRD} simulator in \S\ref{sec:exp-policy-learning} (Fig.~\ref{fig:franka_reach}). The goal of this task is to move the robot’s end-effector to a randomly-specified target position. The robot is controlled via a \textit{joint-position controller}. See the Appendix for detailed task and reward settings. We command the robot to move to 50 different target positions sampled within the robot’s workspace, and we evaluate the policy's performance by measuring the distance to the targets at steady-state. As a baseline, we repeat the same experiment using a policy trained in the ground-truth simulator. Deployment results show that policies trained with both \textit{NeRD} and the ground-truth (\textbf{GT}) simulator achieve low steady-state error, with mean and standard deviation: \textbf{NeRD:} $1.927 \pm 0.699$ \textit{mm}, \textbf{GT:} $4.647 \pm 2.667$ \textit{mm}. We show the distance-to-goal plots of ten executions of \textit{NeRD}-trained policies in Fig.~\ref{fig:franka_reach}. These results validate that the \textit{NeRD} model can effectively learn policies that transfer to the real world.

\subsection{Fine-tunability on Real-World Data: Cube Tossing}
\label{sec:exp-cube-tossing}
\begin{figure*}[t!]
    \centering
    \includegraphics[width=\linewidth]{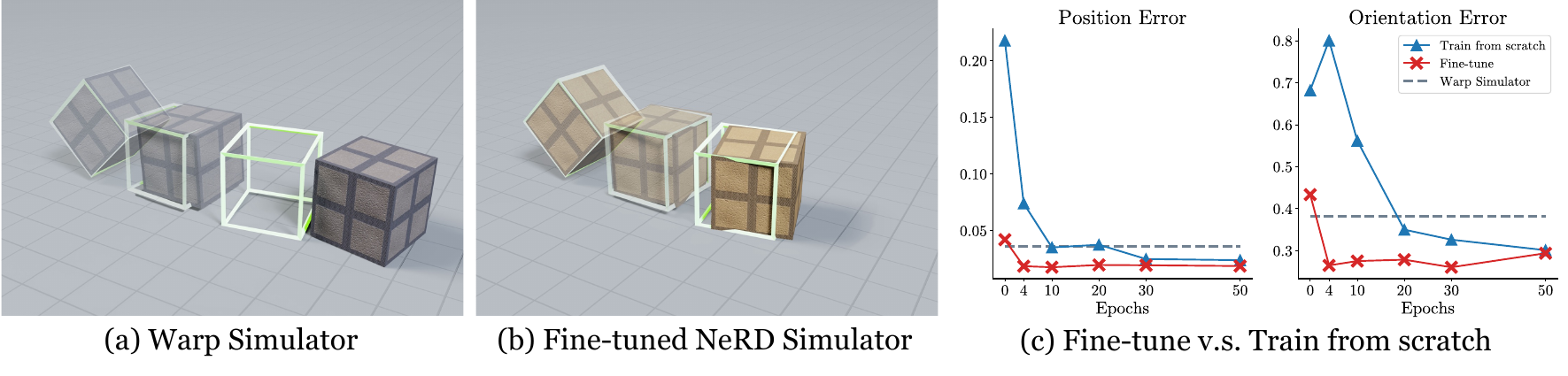}
    
    \caption{\textbf{Fine-tuning of a pretrained \textit{NeRD} model on real-world cube-tossing data.} \textbf{(a-b)} Cube-tossing trajectories simulated by the Warp simulator and by the fine-tuned \textit{NeRD} simulator. The light-green frames are ground-truth poses. \textbf{(c)} Comparison of fine-tuning a pretrained \textit{NeRD} model (red) against training a \textit{NeRD} model from scratch (blue) on the real dataset.}
    \label{fig:cube_toss_comparison}
    \vspace{-1em}
\end{figure*}
We evaluate \textit{NeRD}'s fine-tunability using a real-world cube-tossing dataset \cite{pfrommer2021contactnets}, where a cube is tossed with a random initial state and collides with the ground. We first replicate this cube-tossing environment
in the Warp simulator and generate a synthetic dataset of cube-tossing trajectories for pretraining a \textit{NeRD} model. After pretraining, we fine-tune the \textit{NeRD} model on the real-world dataset. As a comparison, we also train a \textit{NeRD} model from scratch using only the real-world cube-tossing dataset (\ie no simulation data). We provide more details in the Appendix~\ref{sup:cube_toss}. 

We evaluate trained models on $85$ held-out real-world trajectories, and measure the average cube COM position error and orientation error along the trajectory. 
Since the \textbf{Warp simulator} does not fully capture real-world dynamics, it has a position and orientation error of $0.036$ \textit{m} and $0.383$ \textit{rad}, respectively. Both the fine-tuned \textit{NeRD} model and the \textit{NeRD} model trained from scratch outperform Warp. 
Specifically, the \textbf{fine-tuned} \textit{NeRD} model has errors of $0.018$ \textit{m} and $0.266$ \textit{rad}, and the \textbf{model trained from scratch} has errors of $0.023$ \textit{m} and $0.276$ \textit{rad}. Fig. \ref{fig:cube_toss_comparison}(a) and (b) qualitatively compare the trajectories generated by Warp and the fine-tuned \textit{NeRD} model. 
In addition, Fig. \ref{fig:cube_toss_comparison}(c) shows that fine-tuning the pretrained \textit{NeRD} converges in fewer than five training epochs, which is $\bm{10\times}$ \textbf{faster} than training from scratch; thus, pretraining the \textit{NeRD} model on a large-scale simulation dataset enables efficient adaptation to real-world dynamics with a small amount of real-world data.

We further evaluate two baselines designed specifically for this dataset: (1) \textit{GNN-Rigid} \cite{allen2023contact} and (2) \textit{ContactNets} \cite{pfrommer2021contactnets}. For \textit{GNN-Rigid}, we used the 
released model and inference code (training code unavailable). The measured position error is $0.032$ \textit{m}; rotation error is not measured due to missing code. For \textit{ContactNets}, we used the released code to train the model. The evaluated position error is $0.017$ \textit{m} and rotation error is $0.242$ \textit{rad}. These comparisons show that \textit{NeRD} achieves comparable real-world fine-tuning results to specialized models while offering key advantages:  (1) \textit{NeRD} is widely applicable to diverse systems, including articulated and single rigid bodies; (2) Fine-tuning \textit{NeRD} took $<\!10$ min for the \textit{Cube Tossing} dataset, compared to $12$ h for \textit{ContactNets}.


\label{sec:exp-ablation}

\section{Limitations and Future Work}
In this work, we present Neural Robot Dynamics (\textit{NeRD}), learned robot-specific dynamics models capable of stable and accurate simulation over thousands of time steps. Our neural dynamics models can be fine-tuned from real-world data and generalize across different tasks, environments, and low-level controller configurations.

Although our experiments clearly demonstrate the effectiveness of \textit{NeRD}, several promising directions remain for future research. First, while we evaluate \textit{NeRD} on a 14-DoF \textit{Ant} robot and an 18-DoF \textit{ANYmal} robot, we have yet to test it on some of the most complex robotic systems, such as humanoid robots. These advanced robot designs typically have 20-50 degrees of freedom and complex mechanical structures that are difficult to model analytically and simulate efficiently. Applying \textit{NeRD} on these robots can further highlight the efficiency and accuracy benefits of a neural simulation approach. 

Another interesting direction to explore is the trajectory-sampling strategy for generating the synthetic training dataset. Currently, we adopt a random sampling strategy to generate task-agnostic datasets, ensuring the trained \textit{NeRD} model is not limited to specific state distributions. However, random sampling may become ineffective when the state dimensionality grows, particularly for complex robots such as humanoids. Exploring more effective dataset construction strategies that still preserve task-agnostic characteristics of the datasets and the generalizability of the resulting \textit{NeRD} models is a compelling direction for future research. 

Furthermore, our current fine-tuning process assumes we have access to the same state space in the real world as we do in simulation (\ie full robot state, and environment setups for collision detection). However, real-world robot data is often only partially observable due to sensor limitations. Investigating methods to fine-tune a pretrained \textit{NeRD} model from partially observable real-world data is another exciting direction for future research.




\bibliography{references}

\begin{thebibliography}{52}
\providecommand{\natexlab}[1]{#1}
\providecommand{\url}[1]{\texttt{#1}}
\expandafter\ifx\csname urlstyle\endcsname\relax
  \providecommand{\doi}[1]{doi: #1}\else
  \providecommand{\doi}{doi: \begingroup \urlstyle{rm}\Url}\fi

\bibitem[Andrychowicz et~al.(2020)Andrychowicz, Baker, Chociej, Jozefowicz, McGrew, Pachocki, Petron, Plappert, Powell, Ray, et~al.]{andrychowicz2020learning}
O.~M. Andrychowicz, B.~Baker, M.~Chociej, R.~Jozefowicz, B.~McGrew, J.~Pachocki, A.~Petron, M.~Plappert, G.~Powell, A.~Ray, et~al.
\newblock Learning dexterous in-hand manipulation.
\newblock \emph{The International Journal of Robotics Research}, 2020.

\bibitem[Chen et~al.(2023)Chen, Tippur, Wu, Kumar, Adelson, and Agrawal]{chen2023visualdex}
T.~Chen, M.~Tippur, S.~Wu, V.~Kumar, E.~Adelson, and P.~Agrawal.
\newblock Visual dexterity: In-hand reorientation of novel and complex object shapes.
\newblock \emph{Science Robotics}, 2023.

\bibitem[Haarnoja et~al.(2024)Haarnoja, Moran, Lever, Huang, Tirumala, Humplik, Wulfmeier, Tunyasuvunakool, Siegel, Hafner, Bloesch, Hartikainen, Byravan, Hasenclever, Tassa, Sadeghi, Batchelor, Casarini, Saliceti, Game, Sreendra, Patel, Gwira, Huber, Hurley, Nori, Hadsell, and Heess]{Haarnoja2024learning}
T.~Haarnoja, B.~Moran, G.~Lever, S.~H. Huang, D.~Tirumala, J.~Humplik, M.~Wulfmeier, S.~Tunyasuvunakool, N.~Y. Siegel, R.~Hafner, M.~Bloesch, K.~Hartikainen, A.~Byravan, L.~Hasenclever, Y.~Tassa, F.~Sadeghi, N.~Batchelor, F.~Casarini, S.~Saliceti, C.~Game, N.~Sreendra, K.~Patel, M.~Gwira, A.~Huber, N.~Hurley, F.~Nori, R.~Hadsell, and N.~Heess.
\newblock Learning agile soccer skills for a bipedal robot with deep reinforcement learning.
\newblock \emph{Science Robotics}, 2024.

\bibitem[Handa et~al.(2023)Handa, Allshire, Makoviychuk, Petrenko, Singh, Liu, Makoviichuk, Van~Wyk, Zhurkevich, Sundaralingam, et~al.]{handa2023dextreme}
A.~Handa, A.~Allshire, V.~Makoviychuk, A.~Petrenko, R.~Singh, J.~Liu, D.~Makoviichuk, K.~Van~Wyk, A.~Zhurkevich, B.~Sundaralingam, et~al.
\newblock Dextreme: Transfer of agile in-hand manipulation from simulation to reality.
\newblock In \emph{2023 IEEE International Conference on Robotics and Automation (ICRA)}, 2023.

\bibitem[Kumar et~al.(2021)Kumar, Fu, Pathak, and Malik]{kumar2021rma}
A.~Kumar, Z.~Fu, D.~Pathak, and J.~Malik.
\newblock Rma: Rapid motor adaptation for legged robots.
\newblock 2021.

\bibitem[Lee et~al.(2020)Lee, Hwangbo, Wellhausen, Koltun, and Hutter]{lee2020learning}
J.~Lee, J.~Hwangbo, L.~Wellhausen, V.~Koltun, and M.~Hutter.
\newblock Learning quadrupedal locomotion over challenging terrain.
\newblock \emph{Science robotics}, 2020.

\bibitem[Tang et~al.(2024)Tang, Akinola, Xu, Wen, Handa, Van~Wyk, Fox, S.~Sukhatme, Ramos, and Narang]{tang2024automate}
B.~Tang, I.~Akinola, J.~Xu, B.~Wen, A.~Handa, K.~Van~Wyk, D.~Fox, G.~S.~Sukhatme, F.~Ramos, and Y.~Narang.
\newblock Automate: Specialist and generalist assembly policies over diverse geometries.
\newblock In \emph{Robotics: Science and Systems}, 2024.

\bibitem[Funk et~al.(2021)Funk, Schaff, Madan, Yoneda, De~Jesus, Watson, Gordon, Widmaier, Bauer, Srinivasa, et~al.]{funk2021benchmarking}
N.~Funk, C.~Schaff, R.~Madan, T.~Yoneda, J.~U. De~Jesus, J.~Watson, E.~K. Gordon, F.~Widmaier, S.~Bauer, S.~S. Srinivasa, et~al.
\newblock Benchmarking structured policies and policy optimization for real-world dexterous object manipulation.
\newblock \emph{IEEE Robotics and Automation Letters}, 2021.

\bibitem[Gu et~al.(2023)Gu, Xiang, Li, Ling, Liu, Mu, Tang, Tao, Wei, Yao, Yuan, Xie, Huang, Chen, and Su]{gu2023maniskill}
J.~Gu, F.~Xiang, X.~Li, Z.~Ling, X.~Liu, T.~Mu, Y.~Tang, S.~Tao, X.~Wei, Y.~Yao, X.~Yuan, P.~Xie, Z.~Huang, R.~Chen, and H.~Su.
\newblock Maniskill2: A unified benchmark for generalizable manipulation skills.
\newblock In \emph{The Eleventh International Conference on Learning Representations}, 2023.

\bibitem[Li et~al.(2024)Li, Hsu, Gu, Pertsch, Mees, Walke, Fu, Lunawat, Sieh, Kirmani, Levine, Wu, Finn, Su, Vuong, and Xiao]{li24simpler}
X.~Li, K.~Hsu, J.~Gu, K.~Pertsch, O.~Mees, H.~R. Walke, C.~Fu, I.~Lunawat, I.~Sieh, S.~Kirmani, S.~Levine, J.~Wu, C.~Finn, H.~Su, Q.~Vuong, and T.~Xiao.
\newblock Evaluating real-world robot manipulation policies in simulation.
\newblock \emph{arXiv preprint arXiv:2405.05941}, 2024.

\bibitem[Liu et~al.(2021)Liu, Liu, Qin, Xiang, Gou, Xin, Roa, Calli, Su, Sun, et~al.]{liu2021ocrtoc}
Z.~Liu, W.~Liu, Y.~Qin, F.~Xiang, M.~Gou, S.~Xin, M.~A. Roa, B.~Calli, H.~Su, Y.~Sun, et~al.
\newblock Ocrtoc: A cloud-based competition and benchmark for robotic grasping and manipulation.
\newblock \emph{IEEE Robotics and Automation Letters}, 2021.

\bibitem[Du et~al.(2016)Du, Schulz, Zhu, Bickel, and Matusik]{du2016_copter}
T.~Du, A.~Schulz, B.~Zhu, B.~Bickel, and W.~Matusik.
\newblock Computational multicopter design.
\newblock \emph{ACM Trans. Graph.}, 2016.

\bibitem[Li et~al.(2023)Li, Antonova, Sadigh, and Bohg]{li2023tool}
M.~Li, R.~Antonova, D.~Sadigh, and J.~Bohg.
\newblock Learning tool morphology for contact-rich manipulation tasks with differentiable simulation.
\newblock In \emph{2023 IEEE International Conference on Robotics and Automation (ICRA)}, 2023.

\bibitem[Xu et~al.(2021)Xu, Chen, Zlokapa, Foshey, Matusik, Sueda, and Agrawal]{xu2021design}
J.~Xu, T.~Chen, L.~Zlokapa, M.~Foshey, W.~Matusik, S.~Sueda, and P.~Agrawal.
\newblock {An End-to-End Differentiable Framework for Contact-Aware Robot Design}.
\newblock In \emph{Proceedings of Robotics: Science and Systems}, 2021.

\bibitem[Pfrommer et~al.(2021)Pfrommer, Halm, and Posa]{pfrommer2021contactnets}
S.~Pfrommer, M.~Halm, and M.~Posa.
\newblock Contactnets: Learning discontinuous contact dynamics with smooth, implicit representations.
\newblock In \emph{Proceedings of the 2020 Conference on Robot Learning}, 2021.

\bibitem[Hoffman et~al.(2025)Hoffman, Cheng, Li, and Coros]{hoffman2025learning}
B.~Hoffman, J.~Cheng, C.~Li, and S.~Coros.
\newblock Learning more with less: Sample efficient dynamics learning and model-based rl for loco-manipulation.
\newblock \emph{arXiv preprint arXiv:2501.10499}, 2025.

\bibitem[Ajay et~al.(2018)Ajay, Wu, Fazeli, Bauza, Kaelbling, Tenenbaum, and Rodriguez]{ajay2018augmenting}
A.~Ajay, J.~Wu, N.~Fazeli, M.~Bauza, L.~P. Kaelbling, J.~B. Tenenbaum, and A.~Rodriguez.
\newblock Augmenting physical simulators with stochastic neural networks: Case study of planar pushing and bouncing.
\newblock In \emph{2018 IEEE/RSJ International Conference on Intelligent Robots and Systems (IROS)}, 2018.

\bibitem[Zeng et~al.(2020)Zeng, Song, Lee, Rodriguez, and Funkhouser]{zeng2020tossingbot}
A.~Zeng, S.~Song, J.~Lee, A.~Rodriguez, and T.~Funkhouser.
\newblock Tossingbot: Learning to throw arbitrary objects with residual physics.
\newblock \emph{IEEE Transactions on Robotics}, 2020.

\bibitem[Janner et~al.(2019)Janner, Fu, Zhang, and Levine]{janner2019mbpo}
M.~Janner, J.~Fu, M.~Zhang, and S.~Levine.
\newblock When to trust your model: Model-based policy optimization.
\newblock In \emph{Advances in Neural Information Processing Systems}, 2019.

\bibitem[Li et~al.(2025)Li, Krause, and Hutter]{li2025roboticworldmodelneural}
C.~Li, A.~Krause, and M.~Hutter.
\newblock Robotic world model: A neural network simulator for robust policy optimization in robotics, 2025.

\bibitem[Andriluka et~al.(2024)Andriluka, Tabanpour, Freeman, and Sminchisescu]{andriluka2024neural}
M.~Andriluka, B.~Tabanpour, C.~D. Freeman, and C.~Sminchisescu.
\newblock Learned neural physics simulation for articulated 3d human pose reconstruction.
\newblock In \emph{Computer Vision – ECCV 2024: 18th European Conference, Proceedings, Part LXXXIV}, 2024.

\bibitem[Fussell et~al.(2021)Fussell, Bergamin, and Holden]{fussell2021supertrack}
L.~Fussell, K.~Bergamin, and D.~Holden.
\newblock Supertrack: motion tracking for physically simulated characters using supervised learning.
\newblock \emph{ACM Trans. Graph.}, 40\penalty0 (6), 2021.

\bibitem[Hansen et~al.(2022)Hansen, Wang, and Su]{hansen2022temporal}
N.~Hansen, X.~Wang, and H.~Su.
\newblock Temporal difference learning for model predictive control.
\newblock In \emph{International Conference on Machine Learning, PMLR}, 2022.

\bibitem[Macklin(2022)]{warp2022}
M.~Macklin.
\newblock Warp: A high-performance python framework for gpu simulation and graphics.
\newblock \url{https://github.com/nvidia/warp}, 2022.

\bibitem[Bertiche et~al.(2022)Bertiche, Madadi, and Escalera]{bertiche2022cloth}
H.~Bertiche, M.~Madadi, and S.~Escalera.
\newblock Neural cloth simulation.
\newblock \emph{ACM Trans. Graph.}, 2022.

\bibitem[Jin et~al.(2024)Jin, Omens, Geng, Teran, Kumar, Tashiro, and Fedkiw]{jin2024neural}
Y.~Jin, D.~Omens, Z.~Geng, J.~Teran, A.~Kumar, K.~Tashiro, and R.~Fedkiw.
\newblock A neural-network-based approach for loose-fitting clothing.
\newblock \emph{arXiv preprint arXiv:2404.16896}, 2024.

\bibitem[Pfaff et~al.(2021)Pfaff, Fortunato, Sanchez-Gonzalez, and Battaglia]{pfaff2021learning}
T.~Pfaff, M.~Fortunato, A.~Sanchez-Gonzalez, and P.~W. Battaglia.
\newblock Learning mesh-based simulation with graph networks.
\newblock In \emph{International Conference on Learning Representations}, 2021.

\bibitem[Ladick{\`y} et~al.(2015)Ladick{\`y}, Jeong, Solenthaler, Pollefeys, and Gross]{ladicky2015data}
L.~Ladick{\`y}, S.~Jeong, B.~Solenthaler, M.~Pollefeys, and M.~Gross.
\newblock Data-driven fluid simulations using regression forests.
\newblock \emph{ACM Transactions on Graphics (TOG)}, 2015.

\bibitem[Sanchez-Gonzalez et~al.(2020)Sanchez-Gonzalez, Godwin, Pfaff, Ying, Leskovec, and Battaglia]{sanchezgonzalez2020learning}
A.~Sanchez-Gonzalez, J.~Godwin, T.~Pfaff, R.~Ying, J.~Leskovec, and P.~W. Battaglia.
\newblock Learning to simulate complex physics with graph networks.
\newblock In \emph{International Conference on Machine Learning}, 2020.

\bibitem[Li et~al.(2024)Li, Gao, Dai, Li, Hao, and Qin]{li2024mpmnet}
J.~Li, Y.~Gao, J.~Dai, S.~Li, A.~Hao, and H.~Qin.
\newblock Mpmnet: A data-driven mpm framework for dynamic fluid-solid interaction.
\newblock \emph{IEEE Transactions on Visualization and Computer Graphics}, 2024.

\bibitem[Li et~al.(2023)Li, Qiao, Chen, Jatavallabhula, Lin, Jiang, and Gan]{li2023pac}
X.~Li, Y.-L. Qiao, P.~Y. Chen, K.~M. Jatavallabhula, M.~Lin, C.~Jiang, and C.~Gan.
\newblock Pac-nerf: Physics augmented continuum neural radiance fields for geometry-agnostic system identification.
\newblock \emph{International Conference on Learning Representations (ICLR)}, 2023.

\bibitem[Jiang et~al.(2022)Jiang, Sun, and Liu]{jiang2022contact}
Y.~Jiang, J.~Sun, and C.~K. Liu.
\newblock Data-augmented contact model for rigid body simulation.
\newblock In \emph{Proceedings of The 4th Annual Learning for Dynamics and Control Conference}, 2022.

\bibitem[Allen et~al.(2023{\natexlab{a}})Allen, Guevara, Rubanova, Stachenfeld, Sanchez-Gonzalez, Battaglia, and Pfaff]{allen2023contact}
K.~R. Allen, T.~L. Guevara, Y.~Rubanova, K.~Stachenfeld, A.~Sanchez-Gonzalez, P.~Battaglia, and T.~Pfaff.
\newblock Graph network simulators can learn discontinuous, rigid contact dynamics.
\newblock In \emph{Proceedings of The 6th Conference on Robot Learning}, 2023{\natexlab{a}}.

\bibitem[Allen et~al.(2023{\natexlab{b}})Allen, Rubanova, Lopez-Guevara, Whitney, Sanchez-Gonzalez, Battaglia, and Pfaff]{allen2023learning}
K.~R. Allen, Y.~Rubanova, T.~Lopez-Guevara, W.~F. Whitney, A.~Sanchez-Gonzalez, P.~Battaglia, and T.~Pfaff.
\newblock Learning rigid dynamics with face interaction graph networks.
\newblock In \emph{The Eleventh International Conference on Learning Representations}, 2023{\natexlab{b}}.

\bibitem[Hafner et~al.(2020)Hafner, Lillicrap, Ba, and Norouzi]{Hafner2020Dream}
D.~Hafner, T.~Lillicrap, J.~Ba, and M.~Norouzi.
\newblock Dream to control: Learning behaviors by latent imagination.
\newblock In \emph{International Conference on Learning Representations}, 2020.

\bibitem[Hafner et~al.(2023)Hafner, Pasukonis, Ba, and Lillicrap]{hafner2023dreamerv3}
D.~Hafner, J.~Pasukonis, J.~Ba, and T.~Lillicrap.
\newblock Mastering diverse domains through world models.
\newblock \emph{arXiv preprint arXiv:2301.04104}, 2023.

\bibitem[Hansen et~al.(2024)Hansen, Su, and Wang]{hansen2024tdmpc2}
N.~Hansen, H.~Su, and X.~Wang.
\newblock Td-mpc2: Scalable, robust world models for continuous control, 2024.

\bibitem[Moerland et~al.(2023)Moerland, Broekens, Plaat, Jonker, et~al.]{moerland2023mbrl}
T.~M. Moerland, J.~Broekens, A.~Plaat, C.~M. Jonker, et~al.
\newblock Model-based reinforcement learning: A survey.
\newblock \emph{Foundations and Trends{\textregistered} in Machine Learning}, 2023.

\bibitem[Sanchez-Gonzalez et~al.(2018)Sanchez-Gonzalez, Heess, Springenberg, Merel, Riedmiller, Hadsell, and Battaglia]{sanchez2018graph}
A.~Sanchez-Gonzalez, N.~Heess, J.~T. Springenberg, J.~Merel, M.~Riedmiller, R.~Hadsell, and P.~Battaglia.
\newblock Graph networks as learnable physics engines for inference and control.
\newblock In \emph{International conference on machine learning}, 2018.

\bibitem[Raissi et~al.(2019)Raissi, Perdikaris, and Karniadakis]{raissi2019pinn}
M.~Raissi, P.~Perdikaris, and G.~Karniadakis.
\newblock Physics-informed neural networks: A deep learning framework for solving forward and inverse problems involving nonlinear partial differential equations.
\newblock \emph{Journal of Computational Physics}, 2019.

\bibitem[Cuomo et~al.(2022)Cuomo, Di~Cola, Giampaolo, Rozza, Raissi, and Piccialli]{cuomo2022pinn}
S.~Cuomo, V.~S. Di~Cola, F.~Giampaolo, G.~Rozza, M.~Raissi, and F.~Piccialli.
\newblock Scientific machine learning through physics–informed neural networks: Where we are and what’s next.
\newblock 2022.

\bibitem[Heiden et~al.(2021)Heiden, Millard, Coumans, Sheng, and Sukhatme]{heiden2021neuralsim}
E.~Heiden, D.~Millard, E.~Coumans, Y.~Sheng, and G.~S. Sukhatme.
\newblock Neuralsim: Augmenting differentiable simulators with neural networks.
\newblock In \emph{2021 IEEE International Conference on Robotics and Automation (ICRA)}, 2021.

\bibitem[Kim et~al.(2019)Kim, Yoon, Son, and Lee]{kim2019contact}
M.~Kim, J.~Yoon, D.~Son, and D.~Lee.
\newblock Data-driven contact clustering for robot simulation.
\newblock In \emph{2019 International Conference on Robotics and Automation (ICRA)}, 2019.

\bibitem[Son and Kim(2023)]{son2023local}
D.~Son and B.~Kim.
\newblock Local object crop collision network for efficient simulation of non-convex objects in gpu-based simulators.
\newblock \emph{arXiv preprint arXiv:2304.09439}, 2023.

\bibitem[Featherstone(2007)]{feathersone2007}
R.~Featherstone.
\newblock \emph{Rigid Body Dynamics Algorithms}.
\newblock Springer-Verlag, 2007.

\bibitem[Karpathy(2023)]{nanoGPT}
A.~Karpathy.
\newblock {nanoGPT}, 2023.
\newblock URL \url{https://github.com/karpathy/nanoGPT}.

\bibitem[Radford et~al.(2019)Radford, Wu, Child, Luan, Amodei, Sutskever, et~al.]{radford2019language}
A.~Radford, J.~Wu, R.~Child, D.~Luan, D.~Amodei, I.~Sutskever, et~al.
\newblock Language models are unsupervised multitask learners.
\newblock \emph{OpenAI blog}, 2019.

\bibitem[Sutskever et~al.(2014)Sutskever, Vinyals, and Le]{Ilya2014sequence}
I.~Sutskever, O.~Vinyals, and Q.~V. Le.
\newblock Sequence to sequence learning with neural networks.
\newblock In \emph{Proceedings of the 28th International Conference on Neural Information Processing Systems - Volume 2}, 2014.

\bibitem[Shinbrot et~al.(1992)Shinbrot, Grebogi, Wisdom, and Yorke]{shinbrot1992chaos}
T.~Shinbrot, C.~Grebogi, J.~Wisdom, and J.~A. Yorke.
\newblock Chaos in a double pendulum.
\newblock \emph{American Journal of Physics}, 1992.

\bibitem[Hutter et~al.(2016)Hutter, Gehring, Jud, Lauber, Bellicoso, Tsounis, Hwangbo, Bodie, Fankhauser, Bloesch, Diethelm, Bachmann, Melzer, and Hoepflinger]{hutter2016anymal}
M.~Hutter, C.~Gehring, D.~Jud, A.~Lauber, C.~D. Bellicoso, V.~Tsounis, J.~Hwangbo, K.~Bodie, P.~Fankhauser, M.~Bloesch, R.~Diethelm, S.~Bachmann, A.~Melzer, and M.~Hoepflinger.
\newblock Anymal - a highly mobile and dynamic quadrupedal robot.
\newblock In \emph{2016 IEEE/RSJ International Conference on Intelligent Robots and Systems (IROS)}, 2016.

\bibitem[Schulman et~al.(2017)Schulman, Wolski, Dhariwal, Radford, and Klimov]{schulman2017proximal}
J.~Schulman, F.~Wolski, P.~Dhariwal, A.~Radford, and O.~Klimov.
\newblock Proximal policy optimization algorithms.
\newblock \emph{arXiv preprint arXiv:1707.06347}, 2017.

\bibitem[Makoviychuk et~al.(2021)Makoviychuk, Wawrzyniak, Guo, Lu, Storey, Macklin, Hoeller, Rudin, Allshire, Handa, and State]{makoviychuk2021isaac}
V.~Makoviychuk, L.~Wawrzyniak, Y.~Guo, M.~Lu, K.~Storey, M.~Macklin, D.~Hoeller, N.~Rudin, A.~Allshire, A.~Handa, and G.~State.
\newblock Isaac gym: High performance gpu-based physics simulation for robot learning, 2021.

\end{thebibliography}

\newpage

\appendix








\begin{center}
    \Large \textbf{Neural Robot Dynamics: Appendix}
\end{center}

\startcontents[appendix]
\printcontents[appendix]{l}{1}{\section*{Appendix Contents}}
\thispagestyle{empty}
\newpage





\section{Additional \textit{NeRD} Details}
\label{sup:nerd_details}
We provide additional details about \textit{NeRD} that are not covered in the main paper due to space constraints.

\paragraph{Contact-Related Quantities}
We use contact-related quantities $\bm{\mathcal{C}}_t$ as an application-agnostic representation to capture how the surroundings impact the robot's dynamics, without the need to parameterize the whole environment. This is inspired by the dynamics and contact solvers of analytical simulators, as these solvers also use these contact-related quantities to formulate physics equations to evolve the robot dynamics. We construct $\bm{\mathcal{C}}_t = \{\bm{c}_t^i\}$ by reusing the collision detection module in the classical simulator. Specifically, in our implementation, we adopt a GPU-parallelized collision detection algorithm adapted from the one in Warp. For each pre-specified contact point $\bm{p}_0^i$ on the robot, we use the collision detection algorithm to compute its contact event quantities $\bm{c}_t^i = (\bm{p}_0^i, \bm{p}_1^i, \vec{\bm{n}}^i, d^i)$. Here $\bm{p}_1^i$ is the contact point on a non-robot shape, $\vec{\bm{n}}^i$ is the contact normal, and $d^i$ is the contact distance (zero or negative for collisions). We mask a $\bm{c}_t^i$ to be zero if the associated contact distance is larger than a positive threshold $d^i>\xi$, allowing free-space motion while providing robustness to cases where a collision occurs within the timestep. Quantity $\xi$ is a positive value greater than or equal to the \textit{contact thickness} of a geometry (\ie a standard collision-detection parameter in classical simulators). Ideally, $\xi$ should also exceed the allowed maximum displacement of the contact point within a timestep, ensuring that all near-contact events are retained. However, in practice, the choice of $\xi$ is very flexible, and we use a fixed setting of $\xi = \max(4\cdot\textit{contact\_thickness}, 0.1)$ across all our experiments without task-specific tuning.

\paragraph{Robot-Centric State Representation}
Here we provide the detailed calculation for transforming the robot state into robot's base frame. For the robot state $\bm{s}_k$ at time step $k\in[t-h+1, t]$, we transform it into robot's base frame, $\bm{B}_k$, at time step $k$, where $\bm{B}_k=(\bm{x}_k, \bm{R}_k)$. For the robot articulation, we use the reduced coordinate state, which is spatially invariant; thus, we only need to transform the state of the robot base (\ie $\bm{x}_k$, $\bm{R}_k$, and $\bm{\phi}_k$) into the robot's base frame. Therefore, we have $\bm{s}^{\bm{B}_k}_k = (\bm{x}^{\bm{B}_k}_k, \bm{R}^{\bm{B}_k}_k, \bm{q}_k, \bm{\phi}^{\bm{B}_k}_k, \bm{\dot{q}}_k)$, with
\begin{align}
    \bm{x}^{\bm{B}_k}_k &= \bm{0}, \\
    \bm{R}^{\bm{B}_k}_k &= \textit{Identity}, \\
    \bm{\nu^{\bm{B}_k}}_k &= \bm{R}^{-1}_k (\bm{\nu}_k - \bm{x}_k \times \bm{\omega}_k), \\
    \bm{\omega^{\bm{B}_k}}_k &= \bm{R}^{-1}_k \bm{\omega}_k,
\end{align}
where $\bm{\nu}$ and $\bm{\omega}$ are the linear and angular components of a spatial twist $\bm{\phi}$, respectively.

Additionally, the predicted state difference $\Delta \bm{s}_{t+1}$ (\ie the network's output) is expressed in the robot's base frame at time step $t$ instead of $t+1$, \ie $\Delta \bm{s}_{t+1}^{\bm{B}_t} \triangleq \bm{s}_{t+1}^{\bm{B}_t} \ominus \bm{s}_{t}^{\bm{B}_t}$. This is because, when expressed in its own base frame, the state of the robot base $(\bm{x}, \bm{R})$ is always the identity transformation (\ie $\bm{x}_t^{\bm{B}_t} = \bm{x}_{t+1}^{\bm{B}_{t + 1}} = \bm{0}$ and $\bm{R}_t^{\bm{B}_t} = \bm{R}_{t+1}^{\bm{B}_{t+1}} = \mathbf{I}$), which results in zero state changes for these dimensions, so $\Delta \bm{x}_{t+1}^{\bm{B}_{t+1}} = \mathbf{0}$ and $\Delta \bm{R}_{t+1}^{\bm{B}_{t+1}} = \mathbf{0}$. Therefore the learned model cannot predict the motion of the robot base if using $\Delta\bm{s}^{\bm{B}_{t+1}}_{t+1}$ as the prediction target. 

To compute $\bm{s}^{\bm{B}_t}_{t+1} = (\bm{x}^{\bm{B}_t}_{t+1}, \bm{R}^{\bm{B}_t}_{t+1}, \bm{q}_{t+1}, \bm{\phi}^{\bm{B}_t}_{t+1}, \bm{\dot{q}}_{t+1})$, we use the following calculations:
\begin{align}
    \bm{x}^{\bm{B}_t}_{t+1} &= \bm{R}^{-1}_t(\bm{x}_{t+1} - \bm{x}_t), \\
    \bm{R}^{\bm{B}_t}_{t+1} &= \bm{R}^{-1}_t \bm{R}_{t+1}, \\
    \bm{\nu^{\bm{B}_t}}_{t+1} &= \bm{R}^{-1}_t (\bm{\nu}_{t+1} - \bm{x}_t \times \bm{\omega}_{t+1}), \\
    \bm{\omega^{\bm{B}_t}}_{t+1} &= \bm{R}^{-1}_t \bm{\omega}_{t+1},
\end{align}

\paragraph{Multi-Substep Prediction}
To improve the stability of the simulation, a classical simulator often utilizes a smaller timestep (\ie substep) and runs multiple substeps of solver-integration iterations to obtain the actual state of the robot at the next time step, $\bm{s}_{t+1}$ (as shown in Fig.~\ref{fig:overview}(a)). Unlike previous neural simulation works \cite{sanchez2018graph, andriluka2024neural, allen2023contact} that sequentially predict the robot state acceleration at each substep and obtain the robot state at next time step by time integration over substeps, \textit{NeRD} directly predicts the state difference from the current robot state to the state at the next (macro) time step, $\bm{s}_{t+1}$, which might span multiple substeps in the analytical simulator. This design enables us to learn \textit{NeRD} from a finer-grained simulator with smaller substep sizes without sacrificing the efficiency of the learned model at test time. 

\section{Additional Training Details and Hyperparameters}
\label{sup:training_details}
We adopt a lightweight implementation of causal Transformer \cite{nanoGPT,radford2019language}, and repurpose it as a sequential model for robot dynamics. Past robot-centric simulation states are encoded into embeddings via a learnable linear layer, processed through Transformer blocks with self-attention, and then mapped to latent features from which the robot state difference is predicted as the output. 

We use a fixed set of hyperparameters for training \textit{NeRD} across all six robotic systems -- including training hyperparameters and Transformer hyperparameters -- except for the embedding size of the Transformer model. For robots with fewer degrees of freedom, we use a smaller embedding size to enhance training and inference efficiency. The complete hyperparameter settings used in our experiments are provided in Table \ref{tab:hyperparams}.

\begin{table*}[h!]
\caption{Training hyperparameters in the experiments.}
\centering
\label{tab:hyperparams}
\fontsize{9pt}{11.5pt}\selectfont
\setlength{\tabcolsep}{4pt}
\begin{tabular}{l|l|c|c|c|c|c|c}
\hline
\textbf{Robot}   &          & \textbf{Cartpole} & \textbf{Pendulum} & \textbf{Cube Tossing} & \textbf{Franka} & \textbf{Ant} & \textbf{ANYmal} \\ 
\cline{1-8}
\multirow{3}{*}{Training} & history window size $h$ & \multicolumn{6}{c}{10} \\
\cline{2-8}
& batch size & \multicolumn{6}{c}{512} \\
\cline{2-8}
& learning rate & \multicolumn{6}{c}{linear decay from $1e^{-3}$ to $1e^{-4}$} \\
\hline
\multirow{5}{*}{Transformer} & block size & \multicolumn{6}{c}{32} \\
\cline{2-8}
& num layers & \multicolumn{6}{c}{6} \\
\cline{2-8}
& num heads & \multicolumn{6}{c}{12} \\
\cline{2-8}
& input embedding size & \multicolumn{3}{c|}{192} & \multicolumn{3}{c}{384} \\
\cline{2-8}
& dropout & \multicolumn{6}{c}{0} \\
\hline
\multirow{2}{*}{Output MLP} & num layers & \multicolumn{6}{c}{1} \\
\cline{2-8}
& layer size & \multicolumn{6}{c}{64} \\
\hline
\end{tabular}
\end{table*}


\section{Additional Experiment Details}
\label{sup:exp_details}
\subsection{Details of Policy Learning Tasks}
\label{sup:tasks}
We train a \textit{NeRD} model for each robotic system and use the trained \textit{NeRD} model in all the downstream tasks for the corresponding robotic system. Here we provide details of the policy learning tasks in \S\ref{sec:exp-policy-learning}.

\subsubsection{Cartpole}
\paragraph{Swing-Up Task}
In this task, a \textit{Cartpole} (2-DoF) is controlled to swing up its pole from a randomized initial angle to be upright and maintain the upright pose as long as possible. The \textit{Cartpole} is directly controlled by the commanded 1D joint-space torque of the base prismatic joint. The observation of the policy is $4$-dimensional, including:
\begin{itemize}
    \item 2-dim joint positions: $x, \theta$
    \item 2-dim joint velocities: $\dot{x}, \dot{\theta}$
\end{itemize}
A trajectory is terminated if it exceeds the maximum number of steps $300$, or the cart position of the \textit{Cartpole} moves outside the [$-4\,$\textit{m}, $4\,$\textit{m}] range, or the joint velocity is above $10$ \textit{rad/s}.

The stepwise reward function is below:
$$\mathcal{R}_t = 5 - \theta_t ^2 - 0.05 x_t^2 - 0.1 \dot{\theta_t}^2 - 0.1\dot{x_t}^2.$$

\subsubsection{Ant}
\paragraph{Ant Running}
In this task, an \textit{Ant} robot (14-DoF) is controlled to move forward as fast as possible. The action space is 8D joint-space torque of Ant's non-base revolute joints, and the \textit{Ant} is directly controlled by the commanded joint-space torque. The observation space has 29 dimensions, including:
\begin{itemize}
    \item 1-dim height of the base $h$
    \item 4-dim orientation of the base represented by a quaternion
    \item 3-dim linear velocity of the base $\bm{v}$
    \item 3-dim angular velocity of the base $\bm{\omega}$
    \item 8-dim joint positions
    \item 8-dim joint velocities
    \item 2-dim up and heading vector projections: $p_\text{up}, p_\text{heading}$.
\end{itemize}
The episode is terminated if it exceeds the maximum number of steps $500$, or the height of the base $h$ falls below $0.3$ \textit{m} (\ie $h < 0.3$ \textit{m}). 

The stepwise reward function for the running task is defined below:
$$\mathcal{R}_t = \bm{v}_x + 0.1 p_\text{up} + p_\text{heading}.$$

\paragraph{Ant Spinning}
An \textit{Ant} is controlled to maximize its spinning speed around the gravity axis (Y-axis in this environment) in this task. It uses the same observation space and the termination condition as the running task. The stepwise reward function is defined as below:
$$\mathcal{R}_t = \bm{\omega}_y + p_\text{up}.$$

\paragraph{Ant Spin Tracking}
This task requires an \textit{Ant} to track a spinning speed of $5$ \textit{rad/s}. It uses the same observation space and termination condition as the running task. The stepwise reward function is defined below:
$$\mathcal{R}_t = 5\cdot\exp(-(\bm{\omega}_y - 5)^2 - 0.1\bm{\omega}_x^2-0.1\bm{\omega}_z^2) + 0.1 p_\text{up}.$$

\subsubsection{Franka}
\paragraph{End-Effector Reach Task}
The goal of this task is to move the Franka robot's end-effector to a randomly-specified target position. The action space is defined as delta joint positions, which are executed via a \textit{joint-position PD controller} (Note: the \textit{NeRD} model is still predicting using the joint-space torques, which are converted from the target joint positions via the joint-position PD controller). We also conducted another experiment with \textit{joint-torque control} in \S\ref{sup:franka_joint_torque}. The 13-dim observation space consists of the following:
\begin{itemize}
    \item 7-dim joint positions
    \item 3-dim end-effector position
    \item 3-dim target goal position
\end{itemize}

The episode length of this task is $128$. We adopt an exponential reward function from our existing setups, with minimal tuning:
$$\mathcal{R}_t = -d + \frac{1}{\exp(50d) + \exp(-50d) + \epsilon} + \frac{1}{\exp(300d) + \exp(-300d) + \epsilon}, $$

where $d = \|\vec{e}\|$ is the end-effector's distance to the goal position and $\epsilon = 0.0001$.

\subsubsection{ANYmal}

\paragraph{Forward Walk Velocity-Tracking}
In this task, an \textit{ANYmal} robot \cite{hutter2016anymal} (18-DoF) is controlled to track a forward walking speed of $1$ \textit{m/s}. The action space is the target joint positions, and the \textit{ANYmal} is controlled by a \textit{joint-position PD controller}. The observation space is similar to \textit{Ant} tasks and has $37$ dimensions, including the following:
\begin{itemize}
    \item 1-dim height of the base $h$
    \item 4-dim orientation of the base represented by a quaternion
    \item 3-dim linear velocity of the base $\bm{v}$
    \item 3-dim angular velocity of the base $\bm{\omega}$
    \item 12-dim joint positions
    \item 12-dim joint velocities
    \item 2-dim up and heading vector projections: $p_\text{up}, p_\text{heading}$.
\end{itemize}
The episode is terminated if it exceeds the maximum number of steps $1000$, or the height of the base $h < 0.4$ \textit{m}, or the base or the knees hit the ground. We adopt a commonly used reward function \cite{makoviychuk2021isaac} with minimal tuning:
$$\mathcal{R}_t = \exp\Big(-\left((\bm{v}_x - 1)^2 + \bm{v}_z^2\right)\Big) + 0.5 \exp(-\bm{\omega}_y^2) - (0.002\sum \bm{\tau})^2,$$
where $\bm{\tau}$ is the joint-space torque. Note that, in our \textit{ANYmal} environments, $\vec{x}$ is the forward direction, $\vec{z}$ is the sideways direction, and $\vec{y}$ is the upward direction.

\paragraph{Sideways Walk Velocity-Tracking}
This task requires an \textit{ANYmal} robot to track a sideways walking speed of $1$ \textit{m/s}. It uses the same action space, observation space, and termination condition as the \textit{Forward Walk Velocity-Tracking task}. The reward function is below:
$$\mathcal{R}_t = \exp\Big(-\left(\bm{v}_x^2 + (\bm{v}_z - 1)^2\right)\Big) + 0.5 \exp(-\bm{\omega}_y^2) - (0.002\sum \bm{\tau})^2.$$

\subsection{Visualization and Full Results for Double Pendulum with Varying Contact Environments}
\label{sup:pendulum_results}
\begin{figure}[h!]
    \centering
    \includegraphics[width=\linewidth]{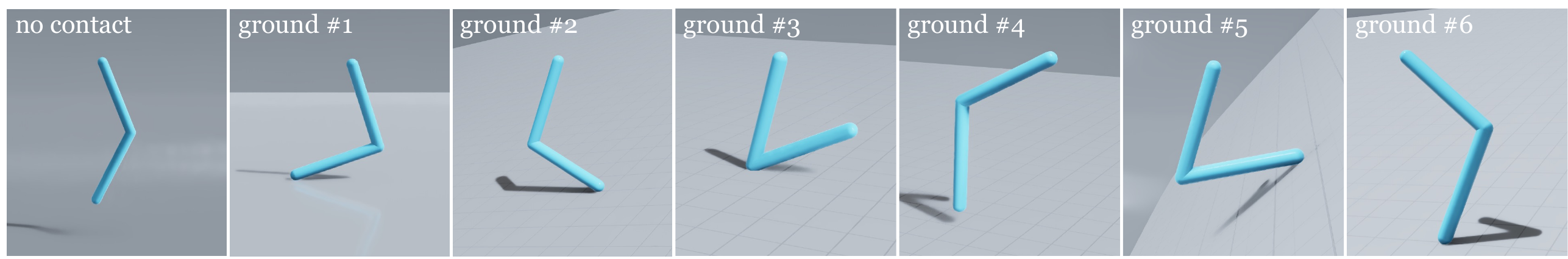}
    
    \caption{\textbf{Seven \textit{Double Pendulum} contact configurations used for testing \textit{NeRD}'s generalizability across different contact environments.}}
    \label{fig:pendulum_with_differnt_grounds_full}
\end{figure}

We provide visualizations of the seven ground configurations and the detailed measured errors in this section.  The seven contact setups include one contact-free scenario where we put the ground far below the pendulum, and six different planar ground settings where the double pendulum is able to make contact with the ground (as shown in Fig.~\ref{fig:pendulum_with_differnt_grounds_full}). We use a single trained \textit{NeRD} model for all contact setups and generate $2048$ passive-motion trajectories of the \textit{Double Pendulum} over a duration of $100$ steps with random initial states and zero joint torques. We report the mean joint-angle errors (in \textit{radians}) over the trajectory in Table \ref{tab:full_pendulum_eval}.

\begin{table*}[h!]
\caption{\textbf{Full passive-motion evaluation results on \textit{Double Pendulum}.} For each contact configuration, we report the mean joint-angle error of each joint in radians.}
\centering
\label{tab:full_pendulum_eval}
\fontsize{8.8pt}{11.5pt}\selectfont
\setlength{\tabcolsep}{4pt}
\begin{tabular}{lccccccc}
\toprule
\textbf{Robot} & \multicolumn{7}{c}{\textbf{Double Pendulum}}  \\ 
\cmidrule(lr){2-8}
Contact Configuration & no contact & ground \#1 & ground \#2 & ground \#3 & ground \#4 & ground \#5 & ground \#6 \\
\midrule
Joint \#1 Error (\textit{rad}) &  $0.004$ & $0.012$ & $0.008$ & $0.005$ & $0.011$ & $0.029$ & $0.008$ \\
Joint \#2 Error (\textit{rad}) & $0.007$ & $0.015$ & $0.011$ & $0.011$ & $0.018$ & $0.056$  & $0.013$ \\
\bottomrule
\end{tabular}
\end{table*}

\subsection{Franka Reach Policy Learning with Joint-Torque Control}
\label{sup:franka_joint_torque}
In \S\ref{sec:exp-policy-learning}, we conduct RL policy-learning experiments to show the \textit{NeRD} model's generalizability to low-level controllers, where we apply different low-level controllers on different robots: \textit{joint-torque control} for \textit{Cartpole} and \textit{Ant}, and \textit{joint-position control} for \textit{Franka} and \textit{ANYmal}. To further verify such generalizability, we conduct another experiment here for the \textit{Franka Reach} task, but with \textit{joint-torque control} instead. We use the same task settings (\eg reward, observation) as \textit{Franka Reach} with \textit{joint-position} control, with the only change being the action space of the policy, and use the same \textit{NeRD} model trained for \textit{Franka}. Similarly, we train three policies with different random seeds \textit{entirely} within the \textit{NeRD} simulator for \textit{Franka}, and then evaluate each learned policy over $2048$ trajectories in the \textit{NeRD} and in the ground-truth simulator and compare the obtained rewards and the standard deviations. Similar to the previous experiments, the results show that the trained \textit{NeRD} model can support high-performing policy learning for different low-level controllers, and the \textit{NeRD} simulator and the ground-truth simulator obtain remarkably similar rewards when evaluating the trained policies (\ie $0.11\%$ error), as reported in Table~\ref{tab:franka_policy_eval}.

\begin{table*}[t!]
\caption{Quantitative evaluation of \textit{Franka Reach} policies trained exclusively in \textit{NeRD} simulators, when deployed in the \textit{NeRD} simulator \textit{or} the ground-truth simulator.}
\centering
\label{tab:franka_policy_eval}
{
\setlength{\tabcolsep}{4pt}
\begin{tabular}{lcc}
\toprule
\textbf{Robot}    &  \multicolumn{2}{c}{\textbf{Franka}} \\ 
\cmidrule(lr){2-3}
\multirow{2}{*}{\textbf{Task}}     &  Reach  & Reach \\
& (joint-position control) & (joint-torque) \\
\midrule
GT Reward         & 89.3 $\pm$ 10.5   & 94.9 $\pm$ 7.8 \\
\cmidrule(lr){2-3}
NeRD Reward       & 91.1 $\pm$ 9.9    & 95.0 $\pm$ 7.8 \\
\cmidrule(lr){2-3}
Reward Err. (\%)  & +2.02\%            & +0.11\%    \\
\bottomrule
\end{tabular}
}
\end{table*}

\subsection{Details of Cube Tossing Experiment Setups}
\label{sup:cube_toss}
We evaluate the fine-tunability of the \textit{NeRD} model using a real-world dataset of cube tossing \cite{pfrommer2021contactnets}, where a cube is tossed with a random initial state and collides with the ground. We first replicate this cube-tossing environment in the Warp simulator by manually tuning the contact and inertia parameters to best replicate the observed dynamics in the dataset. We then generate a dataset comprising $10$K randomly simulated cube-tossing trajectories of length $100$, and pretrain a \textit{NeRD} model from this synthetic dataset. After pretraining, we fine-tune the \textit{NeRD} model on the real-world cube-tossing dataset. The real-world cube-tossing dataset contains $570$ trajectories of varying lengths, corresponding to a total of $60$K dynamics transitions. We split the dataset into $400$ trajectories for training, $85$ trajectories for validation, and $85$ held-out trajectories for testing. To evaluate the fine-tuned model's prediction accuracy, we extract all sub-trajectories of length $80$ (the minimum length of the trajectories in the dataset) from the testing dataset and use the simulator integrated with the fine-tuned \textit{NeRD} model to generate predicted trajectories from the same initial states. We measure the average cube COM position error and average orientation error (in radians) along the trajectory. As a comparison, we also train a \textit{NeRD} model from scratch using only the real-world cube-tossing dataset (\ie no simulation data).

\subsection{Ablation Study}
\label{sup:ablation}
The success of \textit{NeRD} relies on several critical design decisions made during development. In this section, we analyze these design decisions through a series of ablation experiments. Specifically, we conduct our study using two evaluation test cases: (1) contact-free passive motion of \textit{Double Pendulum}; and (2) policy evaluation on the \textit{Ant} running task. All ablation models are trained on the same datasets as the corresponding \textit{NeRD} models. 

For test case \#1, we compute the temporally-averaged mean joint-angle error (average error of two joints) of \textit{Double Pendulum} for each ablation model, from $2048$ passive motion trajectories of the \textit{Double Pendulum} over a duration of $100$ steps with random initial states and zero joint torques. Then we normalize the errors by the error of the \textit{NeRD} model, and report the values in Fig. \ref{fig:ablation} (first row).

For test case \#2, we execute the three \textit{Ant} running policies trained in our policy-learning experiments (\S\ref{sec:exp-policy-learning}) and evaluate the average reward obtained using the simulator integrated with each ablation neural dynamics model ($2048$ trajectories for each policy in each ablation neural dynamics model). For each ablation model, we then compute the reward differences compared to the reward obtained in the ground-truth simulator. The reward differences are then normalized by the reward difference of the \textit{NeRD} model and reported in Fig.~\ref{fig:ablation} (second row).

\subsubsection{Network Architecture}
\begin{figure}[t!]
    \centering
    \includegraphics[width=\linewidth]{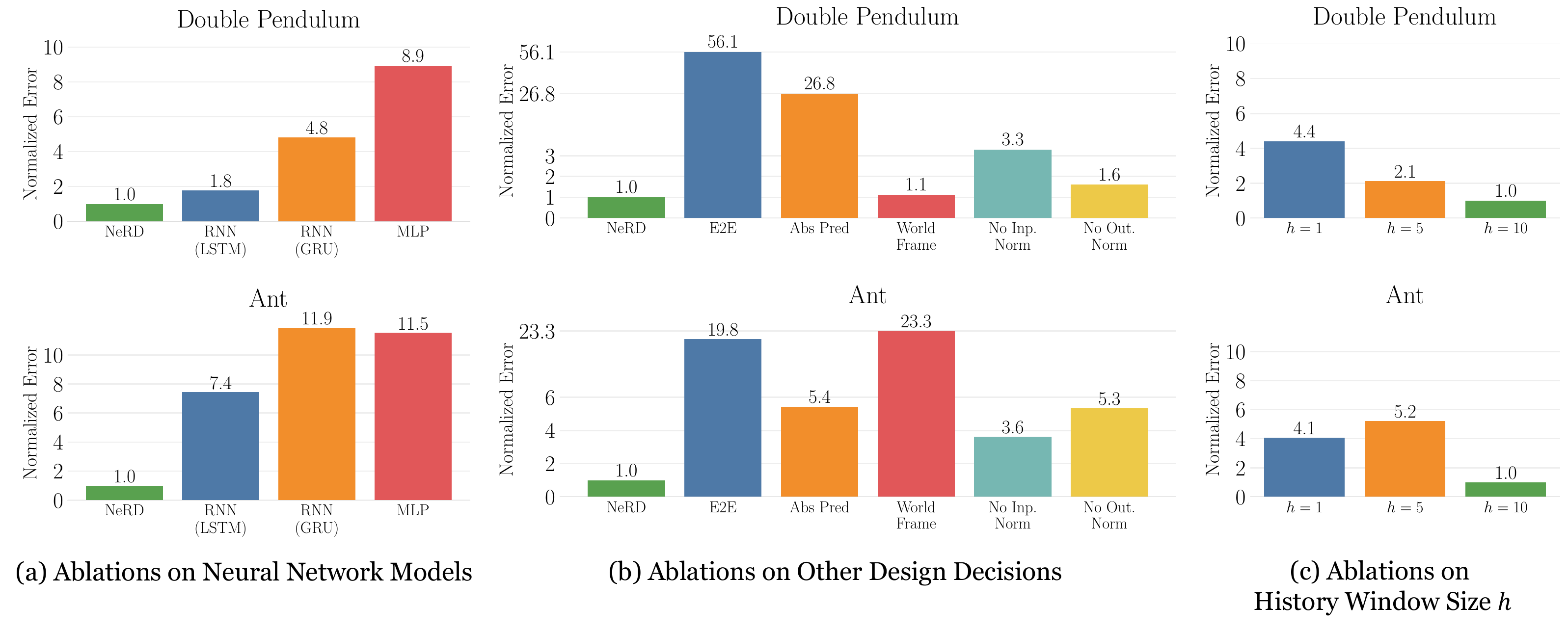}
    
    \caption{\textbf{Ablation Study.} We evaluate ablation variants on two test cases: contact-free passive motion of the \textit{Double Pendulum} and policy evaluation on the \textit{Ant} running task. We normalize the errors by the error of \textit{NeRD} ($h=10$). \textbf{(a)} Ablations of different neural network architectures; \textbf{(b)} Ablations of other critical design decisions in \textit{NeRD}. \textbf{(c)} Ablations on the history window size $h$.}
    \label{fig:ablation}
\end{figure}

During development, we found the Transformer architecture to be the most effective for modeling neural robot dynamics. We demonstrate this by comparing it against three other architectures in Fig.~\ref{fig:ablation}(a): \textbf{MLP}: a baseline model that predicts state changes from the robot-centric simulation state of the current step; \textbf{GRU} and \textbf{LSTM}: two RNN architectures that leverage historical state information in their predictions. Although the ground-truth simulator computes the dynamics in a stateless way (\ie the next state only depends on the current state and torques), we found that sequence modeling is important to achieve high accuracy of the neural robot dynamics model. We hypothesize that the high variance of the velocity inputs is a challenge for the model; by including the historical states as input, the neural model is able to infer a smoothed version of velocity and combine it with the actual velocity input for a better prediction. Furthermore, based on the comparisons for policy evaluation on the \textit{Ant} running task, the causal Transformer model helps achieve a reward that is much closer to the ground-truth simulator, compared to RNN architectures.

\subsubsection{Hybrid Prediction Framework}
To demonstrate the effectiveness of our \textit{Hybrid Prediction Framework}, we compare $\textit{NeRD}$ against an \textit{End-to-End} prediction baseline (\textbf{E2E}), which directly maps robot state and action to the next robot state. This \textit{end-to-end} framework is commonly adopted by prior neural simulators for rigid bodies~\cite{li2025roboticworldmodelneural, fussell2021supertrack,heiden2021neuralsim}. We reimplemented it in the Warp simulator. Specifically, in our implementation, the \textbf{E2E} baseline maps the robot state and the joint torques to the relative next robot state, and the robot state is expressed in the world frame (as an \textit{End-to-End} approach is not aware of contact information and has to rely on world-frame state to track the possible collisions in a fixed environment). We replace the action input commonly used in \textit{End-to-End} approaches with joint-torque input so that we can use the same training dataset as \textit{NeRD} for a fair comparison. Fig. \ref{fig:ablation}(b) shows that the \textbf{E2E} baseline has large prediction errors in both test cases. This is because, in the \textit{Double Pendulum} case, the training dataset consists of varying scenarios of contact configurations. However, the \textbf{E2E} state representation without encoding the environment provides insufficient clues to differentiate distinct contact configurations during training, thus resulting in poor performance. In the \textit{Ant} test case, the \textbf{E2E} baseline fails because the world-frame robot state cannot make a reliable prediction when the \textit{Ant} moves far away from the origin and reaches regions outside the range of the training dataset.

\subsubsection{Relative Robot State Prediction}
The third key design decision is the use of relative robot state prediction, where the model predicts the state difference between the current robot state and the robot state in the next time step, rather than directly predicting the absolute next robot state. Predicting the relative state changes effectively reduces the range of values of the model output, thus stabilizing model training.
We compare \textit{NeRD} against its \textbf{Abs Pred} variant which predicts the absolute next state of the robot, in Fig.~\ref{fig:ablation}(b). The results in the figure show that even for the low-dimensional system like \textit{Double Pendulum}, predicting the absolute state significantly increases training difficulty and results in a prediction error $26\times$ larger than the error of predicting with relative state changes.

\subsubsection{Robot-Centric State Representation}
Next, we design experiments to demonstrate the importance of the robot-centric and spatially-invariant simulation state representation.
For this ablation study, we train a model with the robot state and contact quantities represented in world space (\textbf{World Frame} variant in Fig.~\ref{fig:ablation}(b)), \ie the loss formulation in Eq.~\ref{eq:original_loss}. As shown in the results, using the world-frame representation does not degrade the model's performance in the \textit{Double Pendulum} case of contact-free motion. This is because the pendulum has a revolute base joint that remains fixed in position, limiting the visited states to the domain covered by the training dataset.
In contrast, the world-frame representation fails entirely in the \textit{Ant} running task, as the \textit{Ant} moves far away from the origin during running and quickly reaches regions outside the training dataset's distribution, causing the model to make unreliable predictions.

\subsubsection{Model Input and Output Normalization}
We then show the critical role of normalizing both the inputs and outputs of the neural robot dynamics models. As shown by the \textbf{No Inp. Norm} and \textbf{No Out. Norm} variants in Fig.~\ref{fig:ablation}(b), removing either input or output normalization degrades the performance of the \textit{NeRD} model. This is because the input normalization effectively regularizes the ranges of the inputs to the model, making model training stable and efficient. Meanwhile, output normalization mitigates the dominance of the high-magnitude and high-variance velocity terms in the loss function, balancing prediction accuracy across the different state dimensions.

\subsubsection{History Window Size $h$}
\textit{NeRD} generally gains slight performance improvements when the history window size $h$ increases. We provide a comparison on history window sizes $h=1$, $h=5$, and $h=10$ in Fig. \ref{fig:ablation}(c). We choose $h=10$, as we find $h=10$ consistently provides stable training and generally achieves the best performance across all tasks in our experiments. Though quite rare, we also notice that further increasing $h$ (\eg $h=20$) will occasionally result in an exploded training loss.

\subsection{Computation Speed of \textit{NeRD}}
With 512 parallel Ant envs, Warp (with $16$ substeps, which is the number of substeps we use in Warp for generating training data for Ant) achieves 28K FPS, and \textit{NeRD} achieves 46K FPS. We do not view this comparison as definitive, as both Warp and \textit{NeRD} can be further accelerated; however, as a neural model, \textit{NeRD} benefits from continuous advances in AI hardware and community efforts in ML software acceleration (\textit{e.g.,} TensorRT). Additionally, the policy-learning experiments demonstrate that \textit{NeRD} is fast enough for large-scale on-policy RL.

\end{document}